\documentclass{article}

\usepackage[table]{xcolor}
\usepackage[preprint]{corl_2026} 

\hypersetup{
  pdftitle={Finder: Agentic Closed-Loop Object Finding for Embodied Grounding},
  pdfauthor={Shixiong Xu, Zhiyuan Chen, Song Ding, Rui Luo, Xiaowei Liang, Dongxu Miao, Zhiying Du}
}

\usepackage[utf8]{inputenc}
\usepackage[T1]{fontenc}
\usepackage{url}
\usepackage{booktabs}
\usepackage{multirow}
\usepackage{amsmath}
\usepackage{amsfonts}
\usepackage{nicefrac}
\usepackage{microtype}
\usepackage{enumitem}
\usepackage{graphicx}
\usepackage{wrapfig}
\usepackage{listings}
\usepackage[skins,breakable]{tcolorbox}
\lstdefinestyle{prompt}{
  basicstyle=\scriptsize\ttfamily,
  breaklines=true,
  breakatwhitespace=true,
  columns=fullflexible,
  keepspaces=true,
  frame=none,
  xleftmargin=0pt,
  xrightmargin=0pt
}
\tcbset{
  promptbox/.style={
    enhanced,
    breakable,
    colback=gray!3,
    colframe=black!45,
    coltitle=black,
    fonttitle=\bfseries,
    boxrule=0.4pt,
    arc=1mm,
    left=1mm,
    right=1mm,
    top=1mm,
    bottom=1mm
  }
}

\title{Finder: Agentic Closed-Loop Object Finding for Embodied Grounding}

\author{%
  Shixiong Xu \And Zhiyuan Chen \And Song Ding \And Rui Luo \And Xiaowei Liang \And Dongxu Miao \And Zhiying Du
  \AND
  Xiaomi Robotics
}

\begin{document}
\maketitle


\begin{abstract}
  Finding the object referred to by language in a partially observed 3D scene is a core capability for embodied agents.
  Existing approaches either couple object search with online exploration, which can be costly when relevant observations have already been captured, or query pre-built open-vocabulary maps and scene graphs in a static, one-shot fashion.
  We present \textbf{Finder}, an agentic closed-loop object-finding primitive for embodied grounding.
  Instead of treating grounding as passive retrieval from a fixed scene representation, Finder maintains a typed loop state that links query-conditioned planning, scoped evidence gathering, candidate verification, and \texttt{accept}/\texttt{continue}/\texttt{abort} control.
  When evidence is incomplete or ambiguous, the loop can redirect subsequent perception and comparison rather than simply returning the top retrieved object.
  On open-vocabulary embodied Object Retrieval in Habitat/HM3D and real-world RGB-D scenes, Finder improves the averaged 1m success rate by 15.75 points over strong baselines.
  The same primitive also transfers to sequential object grounding and embodied object-centric question answering, improving spatial and temporal localization without changing the inner grounding protocol.
  Project page: \url{https://finder-vln.github.io}.
\end{abstract}



\section{Introduction}

\begin{figure}[t]
  \centering
  \includegraphics[width=\linewidth]{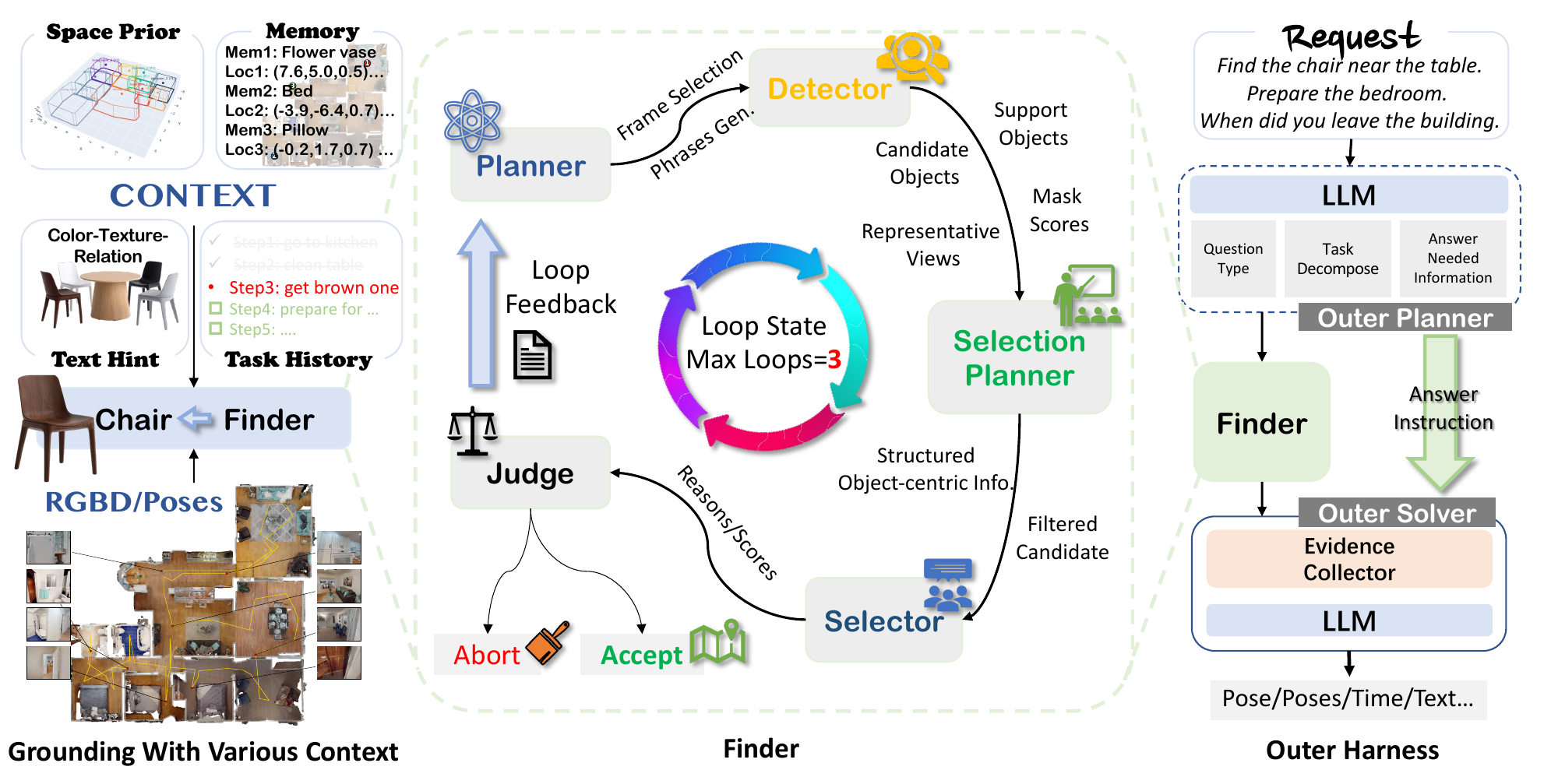}
  \caption{Overview of \textbf{Finder}. Object grounding is a core embodied primitive requiring multi-source context.
  Finder realizes it as a reusable closed-loop inner protocol, wrapped by an outer harness for task- and session-level context injection.}
  \label{fig:teaser}
\end{figure}

For embodied agents, objects are not just entities in a scene; they are anchors for perception, reasoning, and action.
From navigation and manipulation to embodied question answering and long-horizon task execution, many downstream behaviors are organized around identifying, localizing, and revisiting task-relevant objects~\cite{hovsg, rana2023sayplan, alfred, teach, 3dspmr, gorlo2025describe}.
Some systems bind object search to online exploration, which can be expensive when the relevant evidence is already present in captured observations.
Many open-vocabulary systems instead use a representation-centric pipeline: they first build or maintain a semantic map or scene graph from RGB-D observations, and then treat grounding as one-shot retrieval or reranking over that representation~\cite{peng2023openscene, conceptgraphs, hovsg, dualmap, koch2024open3dsg, maggio2024clio, zhang2025open, yang2024llm}.

This paradigm can work well when the target is visually distinctive and the relevant evidence is already present in the stored scene representation.
However, many embodied queries expose three recurring limitations of existing pipelines: resolving \emph{instance ambiguity}, handling \emph{support grounding}, and making \emph{loop-control} decisions explicit.
\textbf{Instance ambiguity} arises when multiple same-category objects are present and the top-ranked match is not necessarily the correct instance~\cite{booker2024embodiedrag,chang2026rag}.
\textbf{Support grounding} matters for relational queries, where supporting objects must be used selectively as disambiguating evidence rather than folded into text matching~\cite{yang2024llm,anwar2025remembr}.
\textbf{Implicit loop control} arises when the current evidence is insufficient and the system must decide what to verify next, where to search next, and whether another round is worthwhile.
In most existing pipelines, these decisions remain implicit, and information from earlier attempts is difficult to carry forward~\cite{inheritsg}.

Therefore, we argue that embodied systems need a reusable agentic closed-loop primitive for object finding.
Such a primitive should not merely score candidates from a fixed scene representation.
Instead, it should support an agentic loop that decides what evidence to gather next in task context (Figure~\ref{fig:teaser}, left).
This abstraction makes grounding both analyzable and reusable, while allowing downstream tasks to inject task- and session-level context without redesigning the inner loop itself.

We instantiate this primitive as \textbf{Finder}, shown in the center of Figure~\ref{fig:teaser},
which organizes one grounding episode as a five-stage loop:
\emph{plan $\rightarrow$ detect $\rightarrow$ selection plan $\rightarrow$ select $\rightarrow$ judge}.
The planner proposes a primary target, optional supporting objects, and a scoped search budget, so the system can decide what to look for and where to gather evidence next.
Candidate-centric context and support-aware verification address instance ambiguity and relational disambiguation, while the final judge determines whether to \emph{accept}, \emph{continue}, or \emph{abort}.
In this way, Finder turns object grounding from one-shot retrieval into a query-conditioned loop over evidence inspection, candidate comparison, and explicit accept/continue/abort control. The same loop can be reused through a lightweight outer harness as shown in Figure~\ref{fig:teaser}: downstream tasks inject resolved-object history, room hints, or benchmark constraints through the request interface, while the inner protocol remains unchanged.

Across Habitat/HM3D~\cite{ramakrishnan2021habitat, yadav2023habitat} and real-world RGB-D retrieval benchmarks, Finder improves averaged 1m success rate by 15.75 points over strong baselines. The same primitive also transfers effectively to sequential object grounding and embodied object-centric question answering~\cite{chang2025ashita, gorlo2025describe}.

In summary, our contributions are:
\begin{itemize}[leftmargin=*, nosep]
  \item Finder, a reusable agentic closed-loop object-finding primitive that makes evidence gathering, candidate comparison, and accept/continue/abort control explicit.
  \item A lightweight outer harness for task- and session-level context management, including resolved-object history, context injection, and optional memory hints, without changing the inner loop.
  \item Extensive experiments demonstrate the effectiveness of Finder on open-vocabulary object retrieval, sequential object grounding, and embodied object-centric question answering.
\end{itemize}

\section{Related work}

\textbf{Open-vocabulary 3D scene understanding.}
Early semantic mapping systems operated over closed-set categories, limiting their applicability to household environments where the space of possible objects is unbounded~\cite{conceptgraphs}.
Vision-language models such as CLIP enabled open-vocabulary 3D representations, including OpenScene~\cite{peng2023openscene}, ConceptGraphs~\cite{conceptgraphs}, HOV-SG~\cite{hovsg}, Hydra~\cite{hughes2022hydra}, and Open3DSG~\cite{koch2024open3dsg}, with OVO~\cite{ovo}, ZING-3D~\cite{zing3d}, and Clio~\cite{maggio2024clio} moving toward online, incremental, or task-driven variants.
Yet these methods still require an exhaustive pre-built or continuously maintained scene representation, which becomes harder under dynamic changes.
DualMap~\cite{dualmap}, OpenIN~\cite{tang2025openin}, Khronos~\cite{schmid2024khronos}, DovSG~\cite{Yan2024DynamicO3}, DynamicGSG~\cite{ge2025dynamicgsg}, and INHerit-SG~\cite{inheritsg} address this with locally updatable object layers or spatio-temporal restructuring, but still rely on hand-designed update rules that are hard to generalize across diverse environments.
Finder departs from this paradigm by using an agentic loop to decide which query-relevant evidence should be inspected next, rather than passively querying a fixed representation or relying on hand-designed update rules.

\textbf{Agentic systems for embodied AI.}
The integration of large language and vision-language models as decision-making agents has gained traction in embodied AI~\cite{salimpour2025towards}.
SayPlan~\cite{rana2023sayplan} and LLM-Grounder~\cite{yang2024llm} ground LLMs in 3D scene graphs for task planning and visual grounding respectively.
SG-Nav~\cite{yin2024sg} prompts LLMs with online scene graphs for zero-shot object navigation.
RoboMemory~\cite{robomemory}, ReMEmbR~\cite{anwar2025remembr}, Mem2Ego~\cite{zhang2025mem2ego}, EmbodiedRAG~\cite{booker2024embodiedrag}, Affordance-RAG~\cite{korekata2026affordance}, MALLVi~\cite{mallvi}, and 3DSPMR~\cite{3dspmr} use memory, retrieval, or multi-agent reasoning to support long-horizon navigation, manipulation, and sequential tasks.
These systems are primarily designed for manipulation or navigation, and do not address task-driven open-vocabulary retrieval with closed-loop evidence refinement.
Finder instead isolates object finding itself as the reusable unit: an explicit inner loop over planning, evidence gathering, candidate comparison, and accept/continue/abort control that can be wrapped by different downstream tasks.

\textbf{Discussion.}
The works above advance open-vocabulary perception, memory-augmented reasoning, and embodied planning, but most still build or maintain a scene graph first and then consume it passively.
Few systems expose target grounding itself as a reusable closed-loop primitive with explicit state, retry logic, and task-conditioned evidence allocation; Finder is designed to close this gap.

\section{Method}


\subsection{Problem Setup and Finder Overview}
\label{sec:overview}

\paragraph{Problem formulation.}
Let $\mathcal{V} = \{(I_t, D_t, P_t)\}_{t=1}^{T}$ denote a stream of RGB-D observations, where $I_t \in \mathbb{R}^{H \times W \times 3}$ is the color image, $D_t \in \mathbb{R}^{H \times W}$ is the depth map, and $P_t \in SE(3)$ is the camera pose at time $t$.
Given a language-conditioned request $q$, optional injected context $m$, and a finite perception budget, Finder iteratively proposes grounded object hypotheses and decides whether to accept the current 3D hypothesis, continue gathering evidence, or abort.

\begin{figure}[t]
  \centering
  \includegraphics[width=\linewidth]{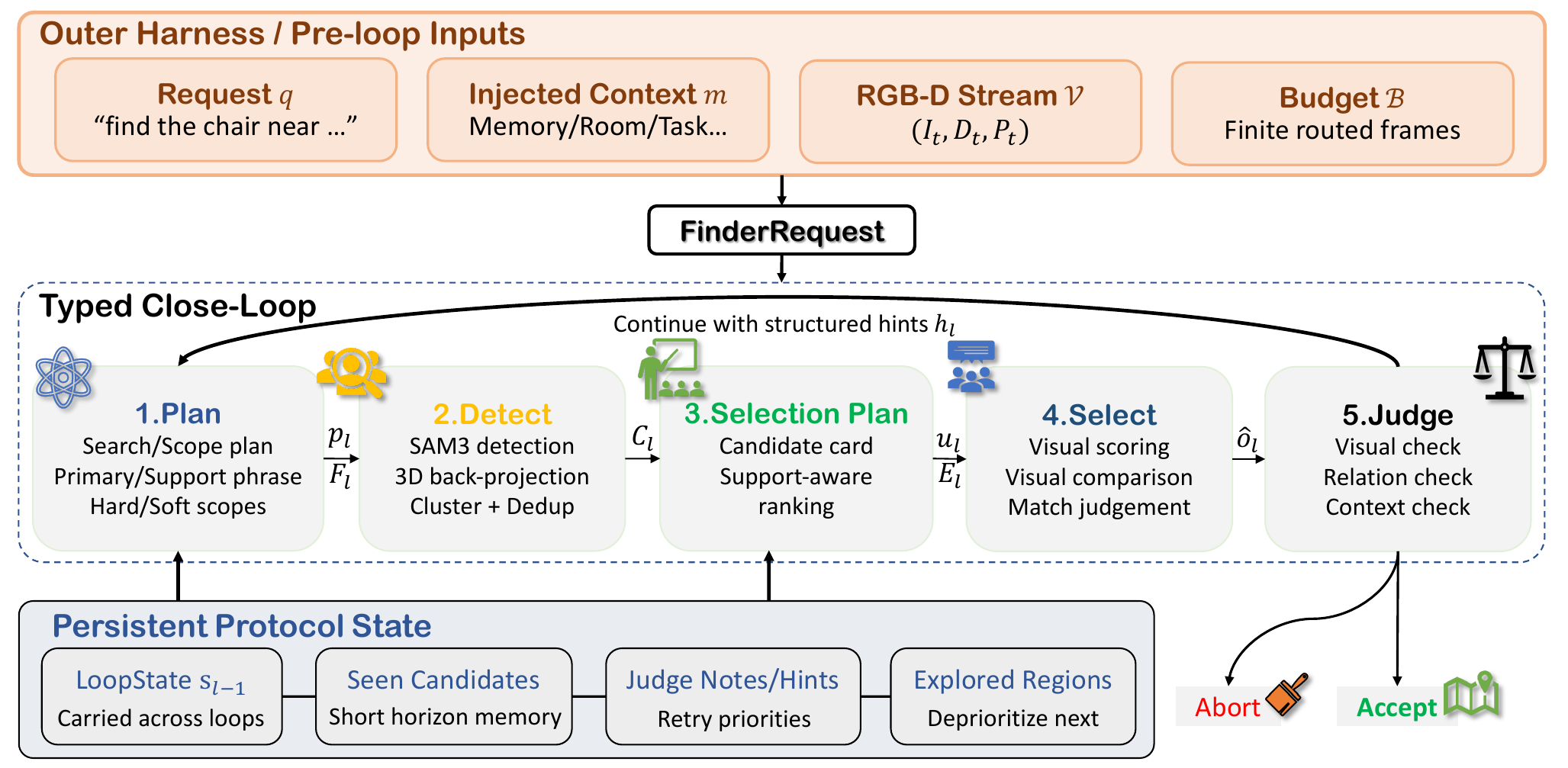}
  \caption{Core execution protocol of \textbf{Finder}. The reusable unit is a typed closed loop: staged execution produces explicit request, plan, detection, context, selection, decision, hint, and snapshot objects, while only the judge can advance the episode to accept, abort, or another planning round.}
  \label{fig:method}
\end{figure}

\paragraph{Closed-loop protocol.}
As shown in Figure~\ref{fig:method}, Finder executes a five-stage closed loop:
\emph{plan $\rightarrow$ detect $\rightarrow$ selection plan $\rightarrow$ select $\rightarrow$ judge}.
Here, request and context injection are treated as pre-loop inputs rather than inner-loop stages.
Within this five-stage view, \emph{plan} includes request understanding, primary/support target extraction, phrase generation, scoped routing, and frame-budget allocation, while \emph{selection plan} includes candidate-context construction and view allocation for verification.
Only the final \emph{judge} stage can terminate the episode or trigger another loop, which makes the stopping boundary explicit and keeps module replacement local.
We summarize one loop as
\begin{align}
(p_{\ell}, F_{\ell}) &= \mathrm{Plan}(q, m, s_{\ell-1}, \mathcal{V}), \\
C_{\ell} &= \mathrm{Detect}(F_{\ell}, p_{\ell}), \\
(u_{\ell}, E_{\ell}) &= \mathrm{SelPlan}(C_{\ell}, p_{\ell}, s_{\ell-1}), \\
\hat{o}_{\ell} &= \mathrm{Select}(u_{\ell}, E_{\ell}), \\
(a_{\ell}, h_{\ell}) &= \mathrm{Judge}(\hat{o}_{\ell}, E_{\ell}, s_{\ell-1}),
\end{align}
where $p_{\ell}$ is the loop search plan, $F_{\ell}$ is the routed frame subset, $C_{\ell}$ is the set of canonicalized candidates, $u_{\ell}$ is the selection plan, $E_{\ell}$ is the candidate-centric evidence context, $a_{\ell} \in \{\texttt{accept}, \texttt{continue}, \texttt{abort}\}$ is the loop action, and $h_{\ell}$ denotes the structured hints carried to the next round.
This protocol makes Finder an iterative evidence-allocation primitive rather than a passive query over a pre-built scene representation.
The protocol defines functional stage boundaries rather than fixing a unique implementation; in our current instantiation, planning and loop judgment are LLM/VLM-driven, while detection is performed by SAM3~\cite{sam3} on the routed frame subset.
The default implementation and API-family sensitivity are reported in Appendix Table~\ref{tab:design_backbone}.

\paragraph{Typed state boundary.}
Finder is reusable because it preserves a stable protocol-state contract across loops and task wrappers.
A \texttt{FinderRequest} stores the query together with injected task/session context and optional memory metadata; a \texttt{SearchPlan} stores the primary/support targets, evidence goals, and scope plan for the current loop; a \texttt{LoopContext} stores candidate evidence and relation summaries for verification; and a \texttt{LoopState} carries seen candidates and judge-produced notes and hints across retries.
This state boundary keeps retries inspectable and lets planners, detectors, selectors, judges, and outer wrappers change without rewriting the closed-loop protocol.

\subsection{Query-Conditioned Planning and Scoped Routing}
\label{sec:planning}

Finder decouples target planning from search control.
A \texttt{SearchPlan} specifies the primary object, optional supporting objects, and compact detection phrases, while a \texttt{ScopePlan} specifies where to search and how to allocate the perception budget.
This separation lets Finder use context to choose where to search while keeping object phrases focused on what to detect.

The frame router executes the scope plan by prioritizing explicit room/floor priors when available, otherwise balancing likely semantic scopes with a global reserve and previously explored constraints.
Supporting targets are also executed adaptively: they can be searched globally, gathered only around primary candidates, or kept as context when active detection would not be useful.
Thus, the planner directs perception toward evidence needed for the current query instead of a full mapping pass.
Additional implementation details are provided in Appendix~\ref{app:planning_evidence_details}.

\subsection{Candidate-Centric Evidence and Verification}
\label{sec:evidence}

Given the routed frame subset and compact phrases, Finder uses phrase-conditioned SAM3~\cite{sam3} detection to form 3D object candidates.
The key design is to keep the subsequent reasoning candidate-centric: instead of flattening all observations into a scene-level summary, Finder builds evidence cards for primary candidates, including representative views, geometric cues, and support-aware relations.

A selection planner then ranks the evidence cards, shortlists candidates, and chooses representative views for direct visual verification.
The VLM selector is restricted to factual image-grounded assessment and match judgments; it does not decide whether the episode should stop.
This keeps visual verification separate from loop control, leaving the judge to make the final \texttt{accept}, \texttt{continue}, or \texttt{abort} decision.
More details on candidate formation and evidence cards are given in Appendix~\ref{app:planning_evidence_details}.

\subsection{Loop Judgment and Cross-Loop Feedback}
\label{sec:judge}
The loop judge is responsible for stopping control: it determines whether the current evidence is sufficient to accept the selected candidate, whether another loop should be executed, or whether the episode should abort.
To do so, it combines four signal families: the selector's visual facts, support-relation metrics, room-binding facts derived from room geometry and candidate pose, and short-horizon history consistency across recent loops.
When the evidence remains insufficient, it emits structured next-loop hints, including candidate follow-up priorities, explored-region constraints, and whether the next loop should favor phrase retargeting or routing changes.
This judge-produced feedback is the mechanism that closes the loop: it separates candidate ranking from stopping control and turns retries into structured follow-up rather than blind repetition.

\subsection{Outer Harness for Task-Specific Context Injection}
\label{sec:harness}
Beyond the default retrieval setting, Finder is adapted to downstream tasks by an outer harness that writes task-specific context into the request interface.
The harness changes the context presented to the loop, but it does not modify the inner loop stages themselves.

\emph{Sequential grounding.}
For sequential grounding, the harness injects the current step label, previously resolved objects, and optionally future-step context.
This lets the same loop ground the next target under task-progress constraints rather than treating each query as an isolated episode.

\emph{Object-centric QA.}
For object-centric question answering, the harness injects already mentioned entities and question-specific constraints.
This lets the same loop retrieve the object evidence needed for answering while preserving the underlying grounding protocol.

\section{Experiments}

\subsection{Experimental Setup}

\paragraph{Evaluation settings.}
We evaluate Finder in three settings: \emph{Object Retrieval}, \emph{Sequential Grounding}, and \emph{Spatio-Temporal QA}.
The first tests the core closed-loop grounding capability, while the latter two test transfer to downstream object-centric embodied tasks.

\emph{Object Retrieval.}
We evaluate closed-loop open-vocabulary retrieval on two splits: a simulated split built on Habitat~\cite{ramakrishnan2021habitat} with HM3D~\cite{yadav2023habitat}, covering \textbf{9} indoor scenes with RGB-D streams and poses, and a real-world split with \textbf{4} cluttered indoor environments from FSR-VLN~\cite{zhou2025fsr}.
Appendices~\ref{app:dataset_annotation} and~\ref{app:experiments} provide dataset construction details, statistics, and split diagnostics.

\emph{Sequential Grounding.}
We evaluate transfer to task-chain grounding on the published SG3D benchmark following ASHiTA~\cite{chang2025ashita} and DAAAM~\cite{gorlo2025describe}.
This setting tests whether the same inner loop can ground an ordered chain of object-level targets rather than a single isolated query.

\emph{Spatio-Temporal QA.}
We evaluate transfer to object-centric question answering on OC-NaVQA following DAAAM~\cite{gorlo2025describe}.
This setting tests whether grounded object evidence can support downstream question answering under different temporal constraints.

\paragraph{Evaluation metrics.}
We evaluate across the three settings summarized in Table~\ref{tab:metrics}.
We use $\hat{\mathbf{x}}, \mathbf{x}^{\star}$ for predicted and ground-truth 3D locations, $\hat{Z}, Z^{\star}$ for predicted and ground-truth step sequences, and $\hat{a}, a^{\star}$, $\hat{\mathbf{p}}, \mathbf{p}^{\star}$, $\hat{t}, t^{\star}$ for predicted and ground-truth QA answers, positions, and times.

\begin{table}[t]
\centering
\caption{Evaluation metrics grouped by task setting.}
\label{tab:metrics}
\small
\begin{tabular*}{\linewidth}{@{\extracolsep{\fill}}l l p{0.58\linewidth}@{}}
\toprule
Setting & Metric & Definition \\
\midrule
\multirow{3}{*}{Object Retrieval}
  & S@0.5 $\uparrow$ & category-correct retrieval rate with $\|\hat{\mathbf{x}}-\mathbf{x}^{\star}\|_2 < 0.5$m \\
  & S@1 $\uparrow$ & category-correct retrieval rate with $\|\hat{\mathbf{x}}-\mathbf{x}^{\star}\|_2 < 1.0$m \\
  & LocErr $\downarrow$ & mean $\|\hat{\mathbf{x}}-\mathbf{x}^{\star}\|_2$ over successful S@1 queries \\
\midrule
\multirow{2}{*}{Sequential Grounding}
  & s-acc $\uparrow$ & fraction of subtasks with predicted pose inside the GT bbox \\
  & t-acc $\uparrow$ & fraction of tasks with all subtask poses inside the GT bboxes \\
\midrule
\multirow{3}{*}{Spatio-Temporal QA}
  & QA Acc. $\uparrow$ & answer match rate, i.e., $\hat{a}=a^{\star}$ \\
  & PosErr $\downarrow$ (m) & mean $\|\hat{\mathbf{p}}-\mathbf{p}^{\star}\|_2$ over questions with spatial supervision \\
  & TempErr $\downarrow$ (min) & mean $|\hat{t}-t^{\star}|$ over questions with temporal supervision \\
\bottomrule
\end{tabular*}
\end{table}


\begin{table}[t]
\centering
\caption{Object Retrieval results on simulated and real-world scenes. S@$r$ denotes category-correct Success@$r$ meters. Err denotes LocErr. Best in \textbf{bold}, second \underline{underlined}.}
\label{tab:retrieval}
\small
\resizebox{0.99\linewidth}{!}{%
\begin{tabular}{l ccc ccc ccc}
\toprule
& \multicolumn{3}{c}{Simulated} & \multicolumn{3}{c}{Real-world} & \multicolumn{3}{c}{Average} \\
\cmidrule(lr){2-4} \cmidrule(lr){5-7} \cmidrule(l){8-10}
Method & S@0.5$\uparrow$ & S@1$\uparrow$ & Err$\downarrow$ & S@0.5$\uparrow$ & S@1$\uparrow$ & Err$\downarrow$ & S@0.5$\uparrow$ & S@1$\uparrow$ & Err$\downarrow$ \\
\midrule
HOV-SG~\cite{hovsg}                 & 33.03 & 50.90 & 0.409 & \underline{31.71} & \underline{42.68} & \underline{0.391} & 32.96 & \underline{50.49} & 0.408 \\
DualMap~\cite{dualmap}              & \underline{38.26} & \underline{51.94} & \underline{0.340} & 9.76 & 19.51 & 0.494 & \underline{36.83} & 50.31 & \underline{0.348} \\
FSR-VLN~\cite{zhou2025fsr}          & 32.65 & 48.90 & 0.419 & 12.20 & 24.39 & 0.532 & 31.62 & 47.67 & 0.422 \\
\midrule
\rowcolor{blue!10}
Finder (Ours)                      & \textbf{61.23} & \textbf{67.29} & \textbf{0.187} & \textbf{35.37} & \textbf{46.34} & \textbf{0.339} & \textbf{59.93} & \textbf{66.24} & \textbf{0.193} \\
\bottomrule
\end{tabular}}
\vspace{-2mm}
\end{table}

\subsection{Main Results}

\paragraph{Object Retrieval.}
As shown in Table~\ref{tab:retrieval}, Finder achieves the best performance in both simulated and real-world settings, and leads on all three averaged Object Retrieval metrics.
Compared with the strongest baseline result on each metric among the three baselines, it improves S@0.5 by \textbf{23.10} points, S@1 by \textbf{15.75} points, and reduces LocErr by \textbf{0.155\,m}.
Performance is consistently higher in simulation than in real-world scenes, which we attribute mainly to real capture artifacts such as motion blur and exposure variation that make these methods less stable.

\begin{table}[t]
\centering
\begin{minipage}[t]{0.40\linewidth}
\centering
\caption{Sequential Grounding results on SG3D~\cite{gorlo2025describe}.}
\label{tab:seq_grounding_main}
\normalsize
\setlength{\tabcolsep}{4pt}
\resizebox{\linewidth}{!}{%
\begin{tabular}{lcc}
\toprule
Method & s-acc $\uparrow$ & t-acc $\uparrow$ \\
\midrule
Hydra+GPT~\cite{hughes2022hydra} & 8.18 & 2.44 \\
Hydra(GT)+GPT~\cite{hughes2022hydra} & 14.2 & 6.34 \\
HOV-SG~\cite{hovsg} & 8.98 & 1.95 \\
ASHiTA~\cite{chang2025ashita} & 21.7 & 8.78 \\
DAAAM+GPT~\cite{gorlo2025describe} & \underline{22.16} & \underline{11.22} \\
\midrule
\rowcolor{blue!10}
Finder & \textbf{25.91} & \textbf{12.20} \\
\bottomrule
\end{tabular}}
\end{minipage}
\hfill
\begin{minipage}[t]{0.57\linewidth}
\centering
\caption{Spatio-Temporal QA results on OC-NaVQA~\cite{gorlo2025describe}.
QA Acc. measures answer correctness, while PosErr and TempErr measure localization quality.}
\label{tab:stqa_main}
\normalsize
\setlength{\tabcolsep}{4pt}
\resizebox{\linewidth}{!}{%
\begin{tabular}{lccc}
\toprule
Method & QA Acc. $\uparrow$ & PosErr $\downarrow$ & TempErr $\downarrow$ \\
\midrule
ReMEmbR-2B~\cite{anwar2025remembr,liu2025nvila} & 0.432 & 53.466 & 2.287 \\
ReMEmbR-8B~\cite{anwar2025remembr,liu2025nvila} & 0.463 & 55.894 & 4.106 \\
ConceptGraphs~\cite{conceptgraphs} & 0.299 & 111.29 & -- \\
DAAAM~\cite{gorlo2025describe} & \textbf{0.711} & \underline{41.75} & \underline{1.792} \\
\midrule
\rowcolor{blue!10}
\textbf{Finder} & \underline{0.540} & \textbf{39.04} & \textbf{1.433} \\
\bottomrule
\end{tabular}}
\end{minipage}
\vspace{-2mm}
\end{table}

\paragraph{Sequential Grounding.}
The results on the SG3D sequential-grounding benchmark~\cite{gorlo2025describe,zhang2024task} are reported in Table~\ref{tab:seq_grounding_main}, which evaluates whether predicted subtask poses fall inside the corresponding ground-truth target bounding boxes across an ordered chain of object-level targets.
Finder achieves the best results on both s-acc and t-acc, improving over DAAAM+GPT~\cite{gorlo2025describe} by \textbf{3.75} and \textbf{0.98} percentage points, respectively.
This is consistent with Finder's strength in precise grounding, since a subtask is counted correct only when its predicted pose falls inside the ground-truth bounding box.

\paragraph{Spatio-Temporal Question Answering.}
The OC-NaVQA results are reported in Table~\ref{tab:stqa_main}.
Finder achieves the best PosErr and TempErr, suggesting that its grounding loop transfers well to spatial and temporal localization questions.
Its QA Acc. remains below that of DAAAM because some questions require aggregation over multiple grounded objects, such as counting bicycles, which is outside Finder's current object-finding focus.

\subsection{Ablation Studies}

We conduct the main ablation on the simulated Object Retrieval split with 1,550 queries; additional implementation ablations are provided in Appendix~\ref{app:api_family}.

\paragraph{Main ablation.}
Table~\ref{tab:ablation_main} isolates the main closed-loop mechanisms by removing or simplifying individual components and comparing support-object execution scopes; detailed variant definitions are listed in Appendix Table~\ref{tab:ablation_variant_defs}.
Four trends are shown in Table~\ref{tab:ablation_main}.
First, iteration without feedback is not enough: \emph{Single-shot} slightly outperforms \emph{No Feedback}, indicating that later loops need carried notes and targeted follow-up rather than blind repetition.
Second, support-aware context matters: \emph{No Support Targets} underperforms the complete support-scope variants, and the global/adaptive/local variants trade off coarse retrieval coverage and fine-grained localization.
Third, rule-based visual decision modules degrade performance, with the larger drop from \emph{Rule Selector} showing that candidate-level visual verification contributes more strongly than the lightweight judge heuristic.
Finally, \emph{LLM Sel.\ Planner} underperforms the default typed selection planner, suggesting that structured shortlist organization is more stable than free-form LLM planning for this stage.

\begin{table}[t]
\centering
\caption{Main ablation on the simulated split. Metrics are reported for simple, hard, and averaged queries; variant definitions are listed in Appendix Table~\ref{tab:ablation_variant_defs}.}
\label{tab:ablation_main}
\footnotesize
\resizebox{\linewidth}{!}{%
\begin{tabular}{l ccc ccc ccc}
\toprule
& \multicolumn{3}{c}{S@0.5$\uparrow$} & \multicolumn{3}{c}{S@1$\uparrow$} & \multicolumn{3}{c}{LocErr$\downarrow$} \\
\cmidrule(lr){2-4} \cmidrule(lr){5-7} \cmidrule(l){8-10}
Variant & Simple & Hard & Avg & Simple & Hard & Avg & Simple & Hard & Avg \\
\midrule
Single-shot          & 60.91 & 55.35 & 58.26 & 66.95 & 59.81 & 63.55 & \textbf{0.187} & \textbf{0.168} & \textbf{0.178} \\
No Feedback          & 59.80 & 55.07 & 57.55 & 66.58 & 59.95 & 63.42 & 0.196 & 0.173 & 0.186 \\
LLM Sel.\ Planner    & 60.79 & 54.67 & 57.87 & 67.32 & 60.22 & 63.94 & \underline{0.192} & 0.178 & 0.186 \\
No Support Targets   & 62.76 & 53.18 & 58.19 & 69.30 & 59.40 & 64.58 & \textbf{0.187} & 0.189 & 0.188 \\
Rule Selector        & 58.20 & 49.80 & 54.19 & 65.72 & 55.89 & 61.03 & 0.204 & 0.195 & 0.200 \\
Rule Judge           & 58.57 & 52.91 & 55.87 & 65.60 & 57.92 & 61.94 & 0.197 & 0.177 & 0.188 \\
\midrule
Finder-Global       & 62.52 & \underline{58.59} & 60.65 & 70.16 & \textbf{65.22} & \textbf{67.81} & 0.195 & 0.183 & 0.189 \\
\rowcolor{blue!10}
Finder-Adaptive     & \textbf{64.12} & 58.05 & \textbf{61.23} & \textbf{70.78} & 63.46 & \underline{67.29} & 0.192 & 0.182 & 0.187 \\
Finder-Local & \underline{63.13} & \textbf{59.00} & \underline{61.16} & \underline{70.41} & \underline{63.73} & 67.23 & 0.196 & \underline{0.169} & \underline{0.184} \\
\bottomrule
\end{tabular}}
\vspace{-2mm}
\end{table}

\begin{figure*}[t]
\centering
\includegraphics[width=0.94\textwidth]{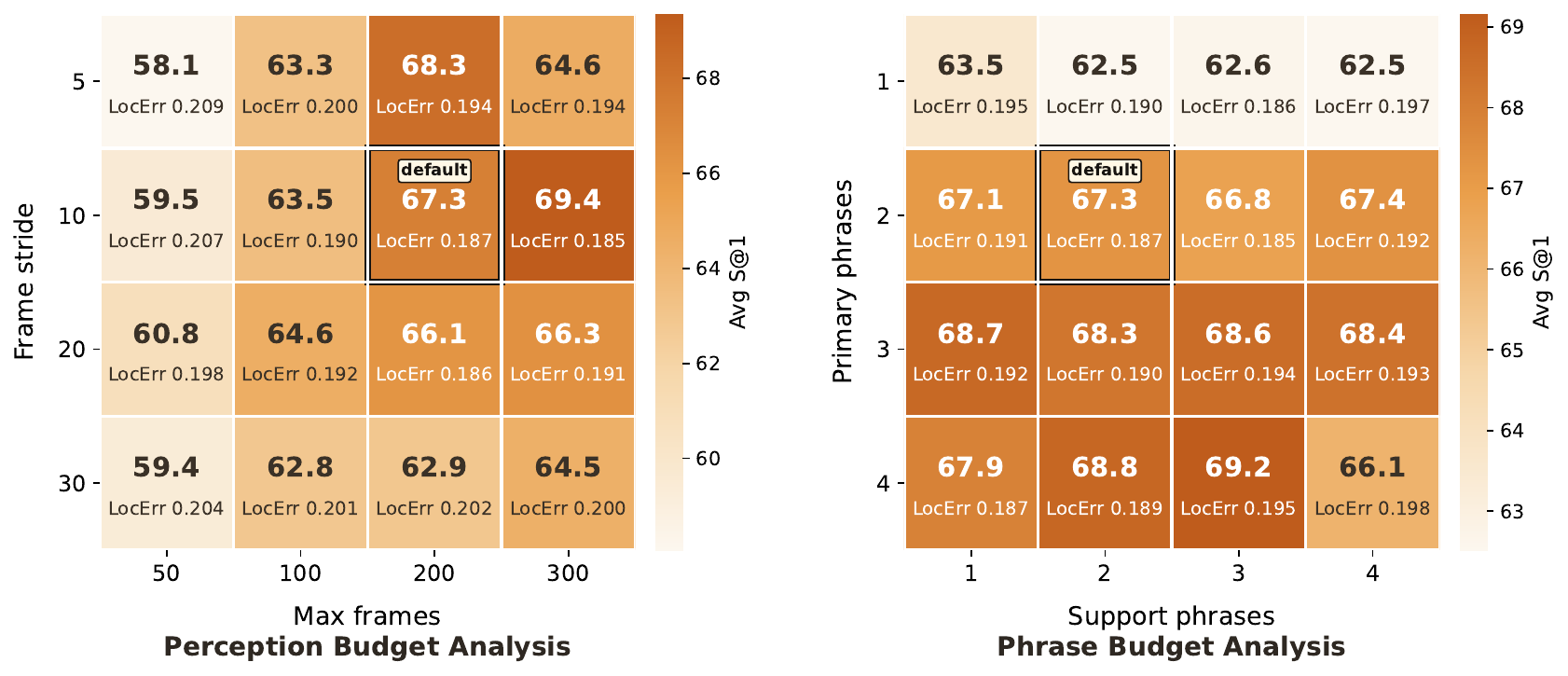}
\caption{Design-space studies for Finder's default budgets. Left: perception budget. Right: phrase budget. The selected defaults balance performance and efficiency.}
\label{fig:design_matrix_preview}
\vspace{-2mm}
\end{figure*}

\paragraph{Perception budget.}
The left panel of Figure~\ref{fig:design_matrix_preview} shows that retrieval improves once the frame budget is large enough to cover relevant views.
However, overly sparse sampling degrades performance because candidate comparison and loop judgment still need sufficiently dense visual evidence.
This supports Finder's default budget choice, which balances evidence coverage and efficiency.

\paragraph{Number of retrieval phrases.}
The right panel of Figure~\ref{fig:design_matrix_preview} shows that compact multi-phrase supervision is useful, especially when moving from a single primary phrase to a small phrase set.
Phrase diversity stabilizes open-vocabulary detection and verification by avoiding dependence on one canonical object wording.
Support phrases provide additional relational evidence, but their gains are less consistent, so the default uses a compact support set rather than maximizing phrase count.

\subsection{Qualitative Analysis}

\begin{figure}[t]
\centering
\includegraphics[width=\linewidth]{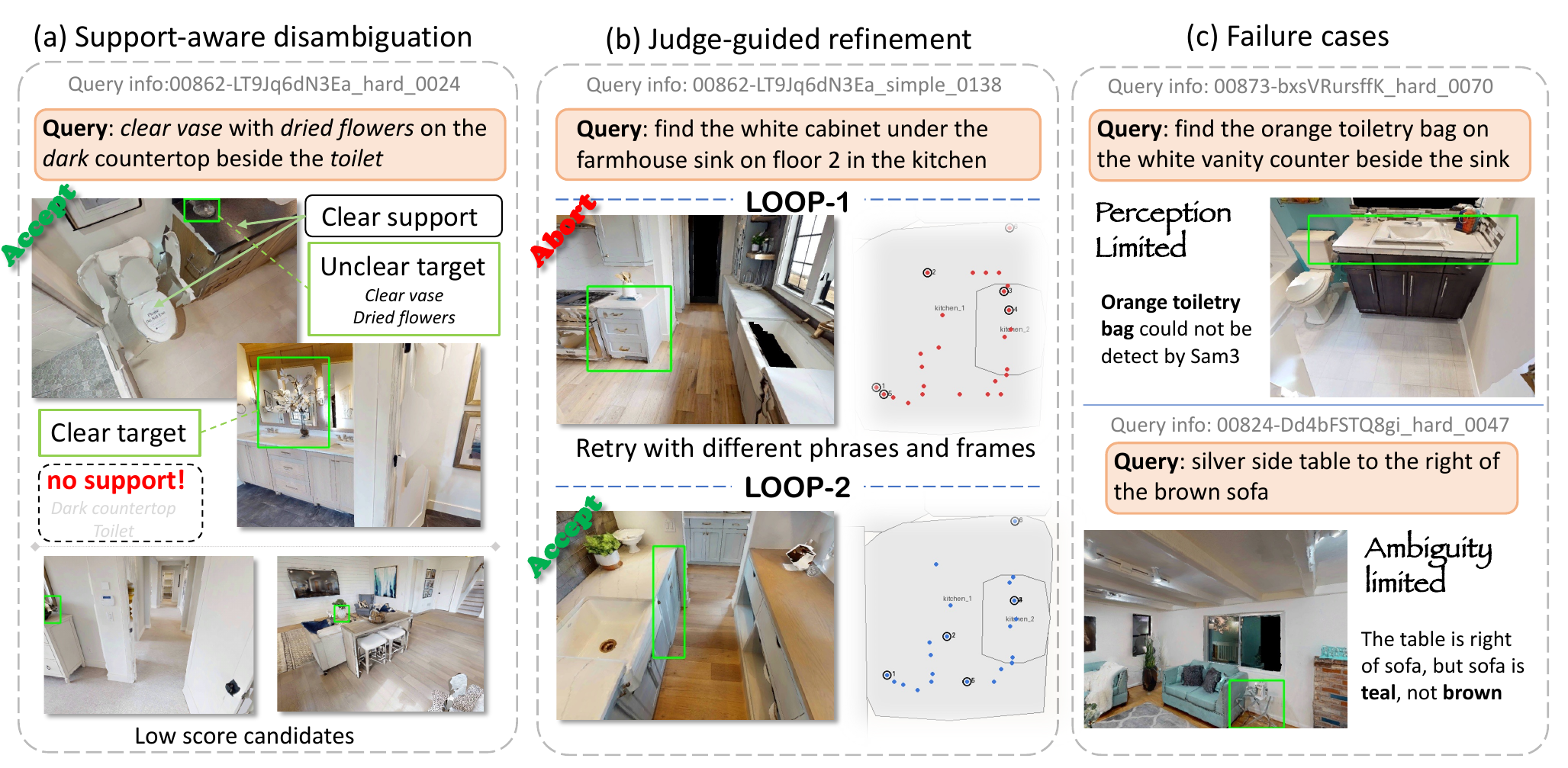}
\caption{Qualitative examples. (a)~Support-aware disambiguation. (b)~Judge-guided cross-loop refinement. (c)~Failure cases: missed target recall and anchor-attribute mismatch.}
\label{fig:qualitative}
\vspace{-2mm}
\end{figure}

\paragraph{Support-aware disambiguation.}
Figure~\ref{fig:qualitative}(a) shows an ambiguous query with multiple visually plausible vase-like candidates.
Finder recovers the correct instance by using the support relation, the vase on the dark countertop beside the toilet, rather than object category alone, rejecting candidates with missing or mismatched local context.

\paragraph{Judge-guided refinement.}
Figure~\ref{fig:qualitative}(b) shows a query where plausible cabinet candidates are retrieved in Loop 1 but the evidence remains incomplete.
Rather than repeating the same search, the judge requests new phrases and frames, shifting Loop 2 toward the sink region and exposing the correct cabinet by filling in evidence missing from the first loop.

\paragraph{Failure cases.}
Figure~\ref{fig:qualitative}(c) highlights two loop-structured failure modes.
The upper example is perception-limited: although the support object is visible and the search region is reasonable, the orange toiletry bag is never proposed as a candidate, leaving later loops with no correct target evidence to refine.
The lower example is ambiguity-limited: Finder retrieves a side table with approximately correct geometry, but the anchor attribute is wrong because the nearby sofa is teal rather than brown.

\section{Conclusion}

We presented Finder, an agentic closed-loop primitive for object-centric grounding.
Finder makes planning, scoped evidence gathering, candidate verification, and accept/continue/abort control explicit under a stable typed loop, while allowing lightweight wrappers to inject task context.
Across object retrieval, sequential grounding, and object-centric question answering, the results show that explicit closed-loop grounding is a practical foundation for broader embodied behavior.

\textbf{Limitations}
Finder currently assumes pre-captured RGB-D streams with known poses; extending the same loop to active exploration and online scene updates would reduce dependence on prior coverage.
The current formulation is strongest for object-finding subproblems and does not yet fully support broader multi-object aggregation or counting.


\clearpage
\acknowledgments{%
}


\bibliography{ref}

\begin{thebibliography}{37}
\providecommand{\natexlab}[1]{#1}
\providecommand{\url}[1]{\texttt{#1}}
\expandafter\ifx\csname urlstyle\endcsname\relax
  \providecommand{\doi}[1]{doi: #1}\else
  \providecommand{\doi}{doi: \begingroup \urlstyle{rm}\Url}\fi

\bibitem[Werby et~al.(2024)Werby, Huang, B{\"u}chner, Valada, and Burgard]{hovsg}
A.~Werby, C.~Huang, M.~B{\"u}chner, A.~Valada, and W.~Burgard.
\newblock Hierarchical open-vocabulary 3d scene graphs for language-grounded robot navigation.
\newblock In \emph{First Workshop on Vision-Language Models for Navigation and Manipulation at ICRA 2024}, 2024.

\bibitem[Rana et~al.(2023)Rana, Haviland, Garg, Abou-Chakra, Reid, and Suenderhauf]{rana2023sayplan}
K.~Rana, J.~Haviland, S.~Garg, J.~Abou-Chakra, I.~Reid, and N.~Suenderhauf.
\newblock Sayplan: Grounding large language models using 3d scene graphs for scalable robot task planning.
\newblock \emph{arXiv preprint arXiv:2307.06135}, 2023.

\bibitem[Shridhar et~al.(2020)Shridhar, Thomason, Gordon, Bisk, Han, Mottaghi, Zettlemoyer, and Fox]{alfred}
M.~Shridhar, J.~Thomason, D.~Gordon, Y.~Bisk, W.~Han, R.~Mottaghi, L.~Zettlemoyer, and D.~Fox.
\newblock {ALFRED}: A benchmark for interpreting grounded instructions for everyday tasks.
\newblock In \emph{Proceedings of the IEEE/CVF Conference on Computer Vision and Pattern Recognition}, pages 10740--10749, 2020.

\bibitem[Padmakumar et~al.(2022)Padmakumar, Thomason, Shrivastava, Lange, Narayan-Chen, Gella, Piramuthu, Tur, and Hakkani-Tur]{teach}
A.~Padmakumar, J.~Thomason, A.~Shrivastava, P.~Lange, A.~Narayan-Chen, S.~Gella, R.~Piramuthu, G.~Tur, and D.~Hakkani-Tur.
\newblock {TEACh}: Task-driven embodied agents that chat.
\newblock In \emph{Proceedings of the AAAI Conference on Artificial Intelligence}, volume~36, pages 2017--2025, 2022.

\bibitem[Cai et~al.(2025)Cai, Du, Wang, and Kong]{3dspmr}
Z.~Cai, Y.~Du, C.~Wang, and Y.~Kong.
\newblock Vision to geometry: 3d spatial memory for sequential embodied mllm reasoning and exploration.
\newblock \emph{arXiv preprint arXiv:2512.02458}, 2025.

\bibitem[Gorlo et~al.(2025)Gorlo, Schmid, and Carlone]{gorlo2025describe}
N.~Gorlo, L.~Schmid, and L.~Carlone.
\newblock Describe anything anywhere at any moment.
\newblock \emph{arXiv preprint arXiv:2512.00565}, 2025.

\bibitem[Peng et~al.(2023)Peng, Genova, Jiang, Tagliasacchi, Pollefeys, Funkhouser, et~al.]{peng2023openscene}
S.~Peng, K.~Genova, C.~Jiang, A.~Tagliasacchi, M.~Pollefeys, T.~Funkhouser, et~al.
\newblock Openscene: 3d scene understanding with open vocabularies.
\newblock In \emph{Proceedings of the IEEE/CVF conference on computer vision and pattern recognition}, pages 815--824, 2023.

\bibitem[Gu et~al.(2024)Gu, Kuwajerwala, Morin, Jatavallabhula, Sen, Agarwal, Rivera, Paul, Ellis, Chellappa, et~al.]{conceptgraphs}
Q.~Gu, A.~Kuwajerwala, S.~Morin, K.~M. Jatavallabhula, B.~Sen, A.~Agarwal, C.~Rivera, W.~Paul, K.~Ellis, R.~Chellappa, et~al.
\newblock Conceptgraphs: Open-vocabulary 3d scene graphs for perception and planning.
\newblock In \emph{2024 IEEE International Conference on Robotics and Automation (ICRA)}, pages 5021--5028. IEEE, 2024.

\bibitem[Jiang et~al.(2025)Jiang, Zhu, Wu, and Song]{dualmap}
J.~Jiang, Y.~Zhu, Z.~Wu, and J.~Song.
\newblock Dualmap: Online open-vocabulary semantic mapping for natural language navigation in dynamic changing scenes.
\newblock \emph{IEEE Robotics and Automation Letters}, 2025.

\bibitem[Koch et~al.(2024)Koch, Vaskevicius, Colosi, Hermosilla, and Ropinski]{koch2024open3dsg}
S.~Koch, N.~Vaskevicius, M.~Colosi, P.~Hermosilla, and T.~Ropinski.
\newblock Open3dsg: Open-vocabulary 3d scene graphs from point clouds with queryable objects and open-set relationships.
\newblock In \emph{Proceedings of the IEEE/CVF Conference on Computer Vision and Pattern Recognition}, pages 14183--14193, 2024.

\bibitem[Maggio et~al.(2024)Maggio, Chang, Hughes, Trang, Griffith, Dougherty, Cristofalo, Schmid, and Carlone]{maggio2024clio}
D.~Maggio, Y.~Chang, N.~Hughes, M.~Trang, D.~Griffith, C.~Dougherty, E.~Cristofalo, L.~Schmid, and L.~Carlone.
\newblock Clio: Real-time task-driven open-set 3d scene graphs.
\newblock \emph{IEEE Robotics and Automation Letters}, 9\penalty0 (10):\penalty0 8921--8928, 2024.

\bibitem[Zhang et~al.(2025)Zhang, Delitzas, Wang, Zhang, Ji, Pollefeys, and Engelmann]{zhang2025open}
C.~Zhang, A.~Delitzas, F.~Wang, R.~Zhang, X.~Ji, M.~Pollefeys, and F.~Engelmann.
\newblock Open-vocabulary functional 3d scene graphs for real-world indoor spaces.
\newblock In \emph{Proceedings of the Computer Vision and Pattern Recognition Conference}, pages 19401--19413, 2025.

\bibitem[Yang et~al.(2024)Yang, Chen, Qian, Madaan, Iyengar, Fouhey, and Chai]{yang2024llm}
J.~Yang, X.~Chen, S.~Qian, N.~Madaan, M.~Iyengar, D.~F. Fouhey, and J.~Chai.
\newblock Llm-grounder: Open-vocabulary 3d visual grounding with large language model as an agent.
\newblock In \emph{2024 IEEE International Conference on Robotics and Automation (ICRA)}, pages 7694--7701. IEEE, 2024.

\bibitem[Booker et~al.(2024)Booker, Byrd, Kemp, Schmidt, and Rivera]{booker2024embodiedrag}
M.~Booker, G.~Byrd, B.~Kemp, A.~Schmidt, and C.~Rivera.
\newblock Embodiedrag: Dynamic 3d scene graph retrieval for efficient and scalable robot task planning.
\newblock \emph{arXiv preprint arXiv:2410.23968}, 2024.

\bibitem[Chang et~al.(2026)Chang, Chen, Zhang, Chen, and Xie]{chang2026rag}
Y.~Chang, R.~Chen, Z.~Zhang, Y.~Chen, and S.~Xie.
\newblock Rag-3dsg: Enhancing 3d scene graphs with re-shot guided retrieval-augmented generation.
\newblock \emph{arXiv preprint arXiv:2601.10168}, 2026.

\bibitem[Anwar et~al.(2025)Anwar, Welsh, Biswas, Pouya, and Chang]{anwar2025remembr}
A.~Anwar, J.~Welsh, J.~Biswas, S.~Pouya, and Y.~Chang.
\newblock Remembr: Building and reasoning over long-horizon spatio-temporal memory for robot navigation.
\newblock In \emph{2025 IEEE International Conference on Robotics and Automation (ICRA)}, pages 2838--2845. IEEE, 2025.

\bibitem[Fang et~al.(2026)Fang, Shi, Qiu, Chen, Shi, Xu, Huo, and Gao]{inheritsg}
Y.~Fang, Z.~Shi, J.~Qiu, Z.~Chen, J.~Shi, H.~Xu, J.~Huo, and Y.~Gao.
\newblock Inherit-sg: Incremental hierarchical semantic scene graphs with rag-style retrieval.
\newblock \emph{arXiv preprint arXiv:2602.12971}, 2026.

\bibitem[Ramakrishnan et~al.(2021)Ramakrishnan, Gokaslan, Wijmans, Maksymets, Clegg, Turner, Undersander, Galuba, Westbury, Chang, et~al.]{ramakrishnan2021habitat}
S.~K. Ramakrishnan, A.~Gokaslan, E.~Wijmans, O.~Maksymets, A.~Clegg, J.~Turner, E.~Undersander, W.~Galuba, A.~Westbury, A.~X. Chang, et~al.
\newblock Habitat-matterport 3d dataset (hm3d): 1000 large-scale 3d environments for embodied ai.
\newblock \emph{arXiv preprint arXiv:2109.08238}, 2021.

\bibitem[Yadav et~al.(2023)Yadav, Ramrakhya, Ramakrishnan, Gervet, Turner, Gokaslan, Maestre, Chang, Batra, Savva, et~al.]{yadav2023habitat}
K.~Yadav, R.~Ramrakhya, S.~K. Ramakrishnan, T.~Gervet, J.~Turner, A.~Gokaslan, N.~Maestre, A.~X. Chang, D.~Batra, M.~Savva, et~al.
\newblock Habitat-matterport 3d semantics dataset.
\newblock In \emph{Proceedings of the IEEE/CVF Conference on Computer Vision and Pattern Recognition}, pages 4927--4936, 2023.

\bibitem[Chang et~al.(2025)Chang, Fermoselle, Ta, Bucher, Carlone, and Wang]{chang2025ashita}
Y.~Chang, L.~Fermoselle, D.~Ta, B.~Bucher, L.~Carlone, and J.~Wang.
\newblock Ashita: Automatic scene-grounded hierarchical task analysis.
\newblock \emph{arXiv preprint arXiv:2504.06553}, 2025.

\bibitem[Hughes et~al.(2022)Hughes, Chang, and Carlone]{hughes2022hydra}
N.~Hughes, Y.~Chang, and L.~Carlone.
\newblock Hydra: A real-time spatial perception system for 3d scene graph construction and optimization.
\newblock \emph{arXiv preprint arXiv:2201.13360}, 2022.

\bibitem[Martins et~al.(2025)Martins, Oswald, and Civera]{ovo}
T.~B. Martins, M.~R. Oswald, and J.~Civera.
\newblock Open-vocabulary online semantic mapping for slam.
\newblock \emph{IEEE Robotics and Automation Letters}, 2025.

\bibitem[Saxena and Chiun(2025)]{zing3d}
P.~Saxena and J.~Chiun.
\newblock Zing-3d: Zero-shot incremental 3d scene graphs via vision-language models.
\newblock \emph{arXiv preprint arXiv:2510.21069}, 2025.

\bibitem[Tang et~al.(2025)Tang, Wang, Deng, Zheng, Deng, Zuo, and Yue]{tang2025openin}
Y.~Tang, M.~Wang, Y.~Deng, Z.~Zheng, J.~Deng, S.~Zuo, and Y.~Yue.
\newblock Openin: Open-vocabulary instance-oriented navigation in dynamic domestic environments.
\newblock \emph{IEEE Robotics and Automation Letters}, 10\penalty0 (9):\penalty0 9256--9263, 2025.

\bibitem[Schmid et~al.(2024)Schmid, Abate, Chang, and Carlone]{schmid2024khronos}
L.~Schmid, M.~Abate, Y.~Chang, and L.~Carlone.
\newblock Khronos: A unified approach for spatio-temporal metric-semantic slam in dynamic environments.
\newblock \emph{arXiv preprint arXiv:2402.13817}, 2024.

\bibitem[Yan et~al.(2025)Yan, Li, Wang, Wu, Wang, Zhu, Chen, and Liu]{Yan2024DynamicO3}
Z.~Yan, S.~Li, Z.~Wang, L.~Wu, H.~Wang, J.~Zhu, L.~Chen, and J.~Liu.
\newblock Dynamic open-vocabulary 3d scene graphs for long-term language-guided mobile manipulation.
\newblock \emph{IEEE Robotics and Automation Letters}, 2025.

\bibitem[Ge et~al.(2025)Ge, Zhu, Yang, and Li]{ge2025dynamicgsg}
L.~Ge, X.~Zhu, Z.~Yang, and X.~Li.
\newblock Dynamicgsg: Dynamic 3d gaussian scene graphs for environment adaptation.
\newblock In \emph{2025 IEEE/RSJ International Conference on Intelligent Robots and Systems (IROS)}, pages 2232--2239. IEEE, 2025.

\bibitem[Salimpour et~al.(2025)Salimpour, Fu, Rachwa{\l}, Bertrand, O'Sullivan, Jakob, Keramat, Militano, Toffetti, Edelman, et~al.]{salimpour2025towards}
S.~Salimpour, L.~Fu, K.~Rachwa{\l}, P.~Bertrand, K.~O'Sullivan, R.~Jakob, F.~Keramat, L.~Militano, G.~Toffetti, H.~Edelman, et~al.
\newblock Towards embodied agentic ai: Review and classification of llm-and vlm-driven robot autonomy and interaction.
\newblock \emph{arXiv preprint arXiv:2508.05294}, 2025.

\bibitem[Yin et~al.(2024)Yin, Xu, Wu, Zhou, and Lu]{yin2024sg}
H.~Yin, X.~Xu, Z.~Wu, J.~Zhou, and J.~Lu.
\newblock Sg-nav: Online 3d scene graph prompting for llm-based zero-shot object navigation.
\newblock \emph{Advances in neural information processing systems}, 37:\penalty0 5285--5307, 2024.

\bibitem[Lei et~al.(2025)Lei, Cai, Cui, Tan, Hong, Hu, Zhu, Wu, Jiang, Wang, et~al.]{robomemory}
M.~Lei, H.~Cai, Z.~Cui, L.~Tan, J.~Hong, G.~Hu, S.~Zhu, Y.~Wu, S.~Jiang, G.~Wang, et~al.
\newblock Robomemory: A brain-inspired multi-memory agentic framework for lifelong learning in physical embodied systems.
\newblock In \emph{NeurIPS 2025 Workshop on Space in Vision, Language, and Embodied AI}, 2025.

\bibitem[Zhang et~al.(2025)Zhang, Liu, Zhang, Aghaei, Hu, Gu, Alomrani, Bravo, Karimi, Hamidizadeh, et~al.]{zhang2025mem2ego}
L.~Zhang, Y.~Liu, Z.~Zhang, M.~Aghaei, Y.~Hu, H.~Gu, M.~A. Alomrani, D.~G.~A. Bravo, R.~Karimi, A.~Hamidizadeh, et~al.
\newblock Mem2ego: Empowering vision-language models with global-to-ego memory for long-horizon embodied navigation.
\newblock \emph{arXiv preprint arXiv:2502.14254}, 2025.

\bibitem[Korekata et~al.(2026)Korekata, Xie, Bisk, and Sugiura]{korekata2026affordance}
R.~Korekata, Q.~Xie, Y.~Bisk, and K.~Sugiura.
\newblock Affordance rag: Hierarchical multimodal retrieval with affordance-aware embodied memory for mobile manipulation.
\newblock \emph{IEEE Robotics and Automation Letters}, 11\penalty0 (3):\penalty0 2706--2713, 2026.

\bibitem[Ahmadi et~al.(2026)Ahmadi, Taji, Kashani, Jadidi, Kashani, and Khalaj]{mallvi}
I.~Ahmadi, M.~Taji, A.~M. Kashani, A.~Jadidi, S.~Kashani, and B.~Khalaj.
\newblock Mallvi: a multi agent framework for integrated generalized robotics manipulation.
\newblock \emph{arXiv preprint arXiv:2602.16898}, 2026.

\bibitem[Carion et~al.(2025)Carion, Gustafson, Hu, Debnath, Hu, Suris, Ryali, Alwala, Khedr, Huang, et~al.]{sam3}
N.~Carion, L.~Gustafson, Y.-T. Hu, S.~Debnath, R.~Hu, D.~Suris, C.~Ryali, K.~V. Alwala, H.~Khedr, A.~Huang, et~al.
\newblock Sam 3: Segment anything with concepts.
\newblock \emph{arXiv preprint arXiv:2511.16719}, 2025.

\bibitem[Zhou et~al.(2025)Zhou, Xiao, Liu, Wang, Chen, Meng, Wang, Feng, Sui, and Su]{zhou2025fsr}
X.~Zhou, T.~Xiao, L.~Liu, Y.~Wang, M.~Chen, X.~Meng, X.~Wang, W.~Feng, W.~Sui, and Z.~Su.
\newblock Fsr-vln: Fast and slow reasoning for vision-language navigation with hierarchical multi-modal scene graph.
\newblock \emph{arXiv preprint arXiv:2509.13733}, 2025.

\bibitem[Liu et~al.(2025)Liu, Zhu, Shi, Zhang, Lou, Yang, Xi, Cao, Gu, Li, et~al.]{liu2025nvila}
Z.~Liu, L.~Zhu, B.~Shi, Z.~Zhang, Y.~Lou, S.~Yang, H.~Xi, S.~Cao, Y.~Gu, D.~Li, et~al.
\newblock Nvila: Efficient frontier visual language models.
\newblock In \emph{Proceedings of the IEEE/CVF Conference on Computer Vision and Pattern Recognition}, pages 4122--4134, 2025.

\bibitem[Zhang et~al.(2024)Zhang, Zhu, Li, Li, Wang, Liu, Ma, Chen, Jia, Huang, et~al.]{zhang2024task}
Z.~Zhang, Z.~Zhu, J.~Li, P.~Li, T.~Wang, T.~Liu, X.~Ma, Y.~Chen, B.~Jia, S.~Huang, et~al.
\newblock Task-oriented sequential grounding and navigation in 3d scenes.
\newblock \emph{arXiv preprint arXiv:2408.04034}, 2024.

\end{thebibliography}

\clearpage
\appendix

\begin{center}
{\Large\bfseries Finder: Agentic Closed-Loop Object Finding for Embodied Grounding\par}
\vspace{0.75em}
{\large\bfseries Appendix\par}
\end{center}
\vspace{1em}

\section{Implementation Details}
\label{app:implementation}

Finder is implemented as a bounded closed loop rather than a one-shot open-vocabulary retrieval call.
The planner first separates object phrases from spatial scope, the router allocates a frame budget under room/floor priors when available, and the judge either accepts the current candidate or emits structured hints for the next loop.
Outer-session memory is optional: previously resolved objects, selected regions, and known context objects enter only as structured planner/judge hints.

\subsection{Default settings}
\label{app:prompts}

Table~\ref{tab:finder_default_settings} lists the default control settings used in the Object Retrieval benchmark.

\begin{table}[h]
\centering
\caption{Default control settings used by Finder in the Object Retrieval benchmark.}
\label{tab:finder_default_settings}
\small
\begin{tabular*}{\linewidth}{@{\extracolsep{\fill}}l l p{0.58\linewidth}@{}}
\toprule
Setting & Default & Role \\
\midrule
Loop budget & 3 loops & Allows retry after zero detections, weak support evidence, or unstable candidates. \\
Frame cap & 200 frames & Caps perception cost after scope-based routing and resampling. \\
Frame stride & 10 & Reduces redundant adjacent views before budgeted detection. \\
Primary phrases & 2 & Provides compact phrase diversity for the target object. \\
Support phrases & 2 & Adds relational/context evidence without dominating the target search. \\
Support execution & Adaptive & Lets the planner choose global, candidate-local, conditional, or no support search. \\
\bottomrule
\end{tabular*}
\end{table}

\subsection{API-family sensitivity}
\label{app:api_family}

Table~\ref{tab:design_backbone} tests whether Finder's typed state contract remains effective under different planner and VLM APIs.
Planner capability is the dominant factor in this axis, while VLM size does not produce a strictly monotonic trend because final performance also depends on planner quality and candidate evidence construction.
We use MiMo-v2-Pro with Qwen 397B-A17B as the default primarily for stable service availability during the evaluation cycle.

\begin{table}[h]
\centering
\caption{API-family ablation on the simulated retrieval split, pairing MiMo planner families with Qwen VLM choices. Cells report S@1 / LocErr; best in \textbf{bold}, second \underline{underlined}.}
\label{tab:design_backbone}
\small
\begingroup
\setlength{\tabcolsep}{5pt}
\begin{tabular}{l cc cc cc}
\toprule
& \multicolumn{2}{c}{35B-A3B} & \multicolumn{2}{c}{122B-A10B} & \multicolumn{2}{c}{397B-A17B} \\
\cmidrule(lr){2-3} \cmidrule(lr){4-5} \cmidrule(l){6-7}
Variant & S@1$\uparrow$ & LocErr$\downarrow$ & S@1$\uparrow$ & LocErr$\downarrow$ & S@1$\uparrow$ & LocErr$\downarrow$ \\
\midrule
\rowcolor{blue!10}
MiMo-v2-Pro / Qwen & 64.71 & 0.194 & 66.65 & \underline{0.188} & \underline{67.29} & \underline{0.187} \\
MiMo-v2.5 / Qwen & \underline{66.06} & \underline{0.190} & \underline{66.90} & \textbf{0.184} & 67.16 & 0.193 \\
MiMo-v2.5-Pro / Qwen & \textbf{68.69} & 0.191 & \textbf{67.81} & 0.190 & 66.97 & \textbf{0.186} \\
Matched Qwen / Qwen & 65.10 & \textbf{0.189} & 63.10 & 0.189 & \textbf{68.00} & 0.193 \\
\bottomrule
\end{tabular}
\endgroup
\end{table}

\subsection{Planning and evidence construction details}
\label{app:planning_evidence_details}

\paragraph{Planning fields.}
The target-side \texttt{SearchPlan} contains one primary target, zero or more supporting targets, an evidence-goal list, and compact phrase sets for targets scheduled for active detection.
The primary target is the object to be returned if the loop succeeds.
Supporting targets are introduced only when they help anchor or disambiguate the primary target, such as in relational queries.
The \texttt{ScopePlan} contains four search-control fields: hard scopes from explicit room/floor constraints, soft scopes from semantic room hints or support cues, per-scope frame budgets with a global reserve, and explored constraints that deprioritize regions or points in later loops.

\paragraph{Routing and support execution.}
If an explicit room or floor is available, Finder spends the budget inside that hard scope.
Otherwise, it splits the budget between likely room scopes and a global reserve.
The frame router applies room/floor filtering, excludes previously explored regions when requested, and pose-samples the remaining frame inventory to satisfy the scope budget.
Each supporting target carries an execution policy: it can be detected globally, deferred until primary candidates are available, or kept as context only.
The local-after-primary mode gathers support-object evidence only on candidate-local views, which is useful when support objects mainly disambiguate among already plausible primary candidates.

\paragraph{Candidate formation and evidence cards.}
For each routed frame, phrase-conditioned masks are backprojected into 3D, clustered into raw object proposals, and merged by geometry-aware deduplication when proposals likely correspond to the same object.
Each retained candidate stores a 3D center, a compact geometry summary, representative views, and provenance linking it to detector instances.
Finder then constructs one evidence card per primary candidate.
Evidence cards summarize candidate-local measurements, visual/geometry cues, support proximity, and the source of support evidence, giving the selector and judge a target-centric record rather than a flattened scene summary.

\subsection{Baseline and variant definitions}
\label{app:baseline_implementations}

All methods are evaluated with the same query splits, ground-truth centroids, and radius-based success metrics defined in Section~\ref{app:retrieval_protocol}.
The baseline implementations are summarized as follows:
\begin{itemize}[leftmargin=*]
    \item \textbf{HOV-SG}~\cite{hovsg}: a pre-built open-vocabulary scene graph baseline. We use top-1 graph retrieval and score the returned object centroid with the same geometry-only success metric as Finder. Hard queries are evaluated from text only, while simple queries additionally use the available room prior for filtering.
    \item \textbf{DualMap}~\cite{dualmap}: a dual semantic-spatial map baseline evaluated under the same hard/simple query protocol. Simulated scenes use the dense-frame rerun results, and real-world scenes are scored after alignment to the shared z-up coordinate convention.
    \item \textbf{FSR-VLN}~\cite{zhou2025fsr}: a navigation-oriented scene-graph/memory baseline adapted to the Object Retrieval queries. We use the same benchmark queries and available simple-query priors, and report the radius-specific retrieval fields when available.
\end{itemize}
For Sequential Grounding and Spatio-Temporal QA, we compare against published benchmark results from ASHiTA~\cite{chang2025ashita}, DAAAM~\cite{gorlo2025describe}, ReMEmbR~\cite{anwar2025remembr}, and ConceptGraphs~\cite{conceptgraphs}, rather than introducing additional task-specific reruns.

\begin{table}[h]
\centering
\caption{Variant-to-change mapping for the main ablation in Table~\ref{tab:ablation_main}.}
\label{tab:ablation_variant_defs}
\small
\begin{tabular*}{\linewidth}{@{\extracolsep{\fill}}l p{0.68\linewidth}@{}}
\toprule
Variant & Change \\
\midrule
Single-shot        & Set \texttt{max\_loops=1}. \\
No Feedback        & Clear carried notes and next-loop hints between loops. \\
LLM Sel.\ Planner  & Replace the default structured selection planner with an LLM-based planner. \\
No Support Targets & Remove supporting targets and candidate-local support follow-up. \\
Rule Selector      & Replace VLM visual verification with a simple heuristic selector. \\
Rule Judge         & Replace the VLM loop judge with a lightweight accept/continue heuristic. \\
Finder-Global      & Force all supporting targets into the global pass. \\
Finder-Adaptive    & Default Finder configuration with planner-chosen support execution. \\
Finder-Local       & Force supporting targets to run in \texttt{local\_after\_primary}. \\
\bottomrule
\end{tabular*}
\end{table}

\section{Dataset Details}
\label{app:dataset_annotation}

The Object Retrieval benchmark contains 1,632 single-object queries over 197 query-target categories across 9 simulated scenes and 4 real-world scenes.
The simulated split is built from RGB-D streams, semantic object annotations, and posed camera trajectories in Habitat/HM3D scenes.
We canonicalize object categories, remove structural or unreachable labels, keep visible instances with valid centroids and view evidence, and export room/floor metadata only as optional priors for scoped simple queries.
The real-world split is built from cluttered indoor RGB-D captures where targets are manually selected from RGB-D frames and lifted to 3D centroids from annotated masks.

\subsection{Dataset statistics}
\label{app:dataset_statistics}

\begin{figure}[h]
  \centering
  \includegraphics[width=\linewidth]{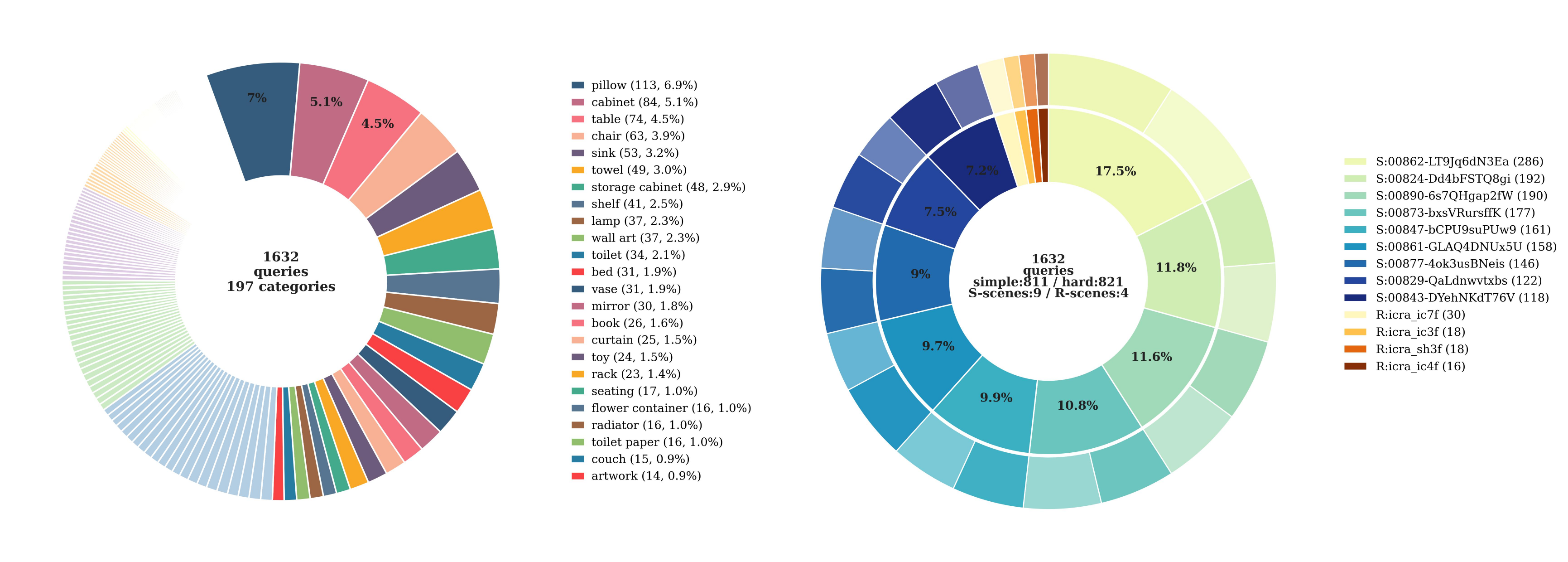}
  \caption{Object Retrieval dataset overview. Left: query distribution across object categories, showing a long-tailed benchmark with 1,632 queries over 197 categories. Right: per-scene query counts and the simple/hard split, with 811 simple and 821 hard queries across 9 simulated and 4 real-world scenes.}
  \label{fig:dataset_stats_main}
\end{figure}

\begin{table}[h]
\centering
\caption{Dataset statistics for the Object Retrieval splits. Object counts refer to retrievable target objects after filtering. Category counts follow the query-target category canonicalization used throughout the benchmark.}
\label{tab:dataset_stats}
\small
\begin{tabular*}{0.88\linewidth}{@{\extracolsep{\fill}}l cc@{}}
\toprule
Statistic & Simulated & Real-world \\
\midrule
Number of scenes                    & 9  & 4 \\
Total rooms                         & 122 & 4 \\
Avg.\ rooms per scene               & 13.6 & 1.0 \\
Total target objects                & 1,890 & 82 \\
Avg.\ target objects per scene      & 210.0 & 20.5 \\
Indexed RGB-D views                 & 31,520 & 67 \\
Avg.\ indexed RGB-D views per scene & 3,502.2 & 16.8 \\
Simple queries                      & 811 & 0 \\
Hard queries                        & 739 & 82 \\
Single-object queries               & 1,550 & 82 \\
Sequential tasks                    & 110 & 0 \\
Query-target categories             & 166 & 49 \\
\bottomrule
\end{tabular*}
\end{table}

\subsection{Annotation workflow and prompts}
\label{app:annotation_pipeline}

Simulated annotations follow four steps: scene inventory, view indexing, query writing, and ground-truth binding.
Simple queries include explicit room/floor priors when available, while hard queries are evaluated scene-wide and must rely on visual, relational, or contextual language rather than hidden room metadata.
Real-world annotations use the same final schema after the selected RGB-D mask is lifted to a 3D object point cloud.

\begin{table}[h]
\centering
\caption{Query-writing prompt constraints used during dataset construction.}
\label{tab:annotation_prompt_summary}
\small
\begin{tabular*}{\linewidth}{@{\extracolsep{\fill}}l p{0.72\linewidth}@{}}
\toprule
Field & Prompt Constraint \\
\midrule
Target binding & Each query must refer to one annotated target object with a canonical category and 3D centroid. \\
Simple query & Include available room/floor scope when it is part of the intended prior. \\
Hard query & Do not reveal room/floor metadata; use visible appearance, relation, or context to identify the target. \\
Ambiguity control & If multiple same-category objects exist, wording should preserve the intended ambiguity level without becoming ungrounded. \\
Quality control & Reject queries whose target is not visible in the evidence views or whose wording cannot be checked from visual context. \\
\bottomrule
\end{tabular*}
\end{table}

\subsection{Annotation interface}
\label{app:annotation_interface}

\begin{figure}[h]
\centering
\begin{minipage}{0.48\linewidth}
\centering
\includegraphics[width=\linewidth]{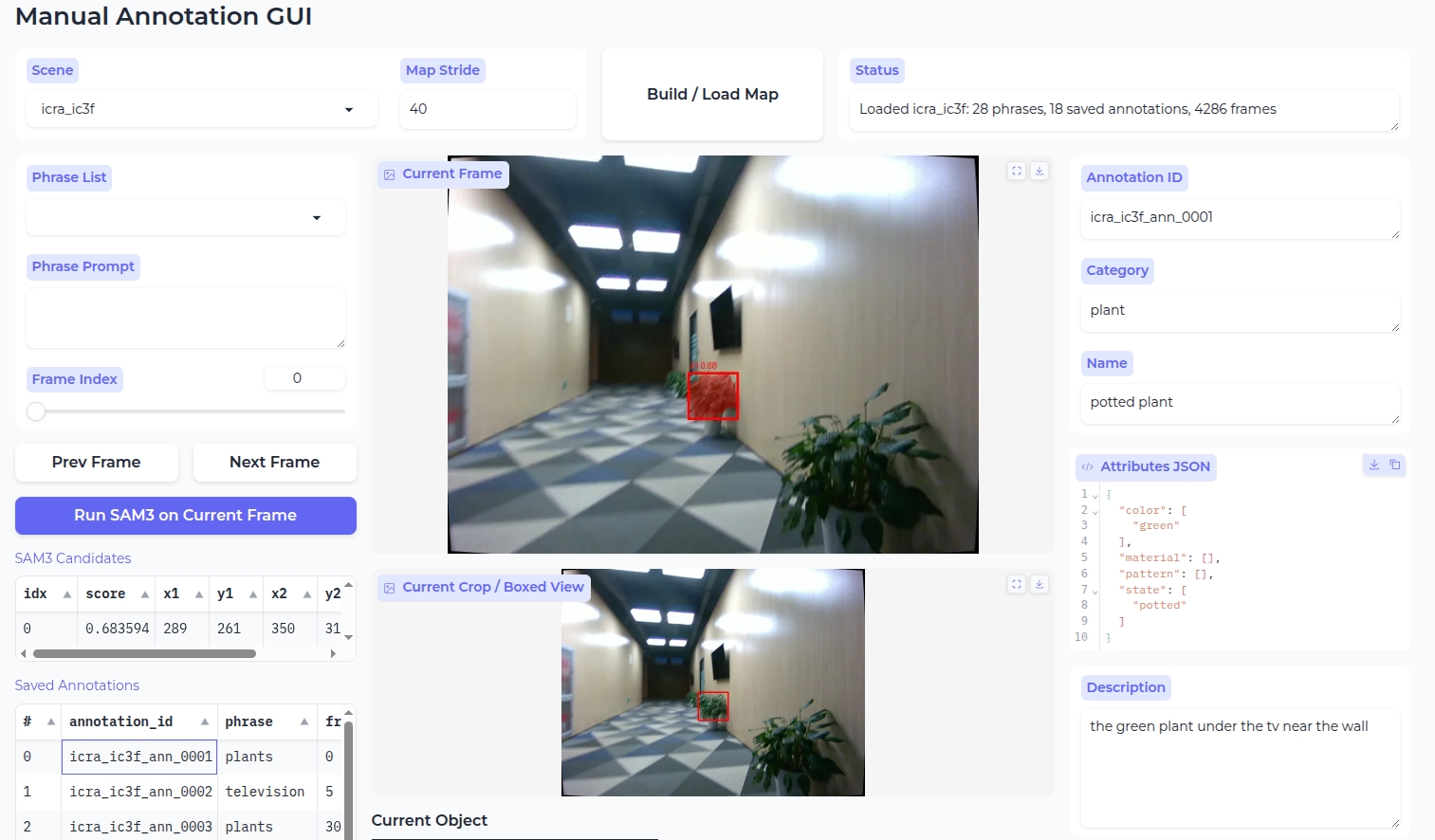}
\end{minipage}
\hfill
\begin{minipage}{0.48\linewidth}
\centering
\includegraphics[width=\linewidth]{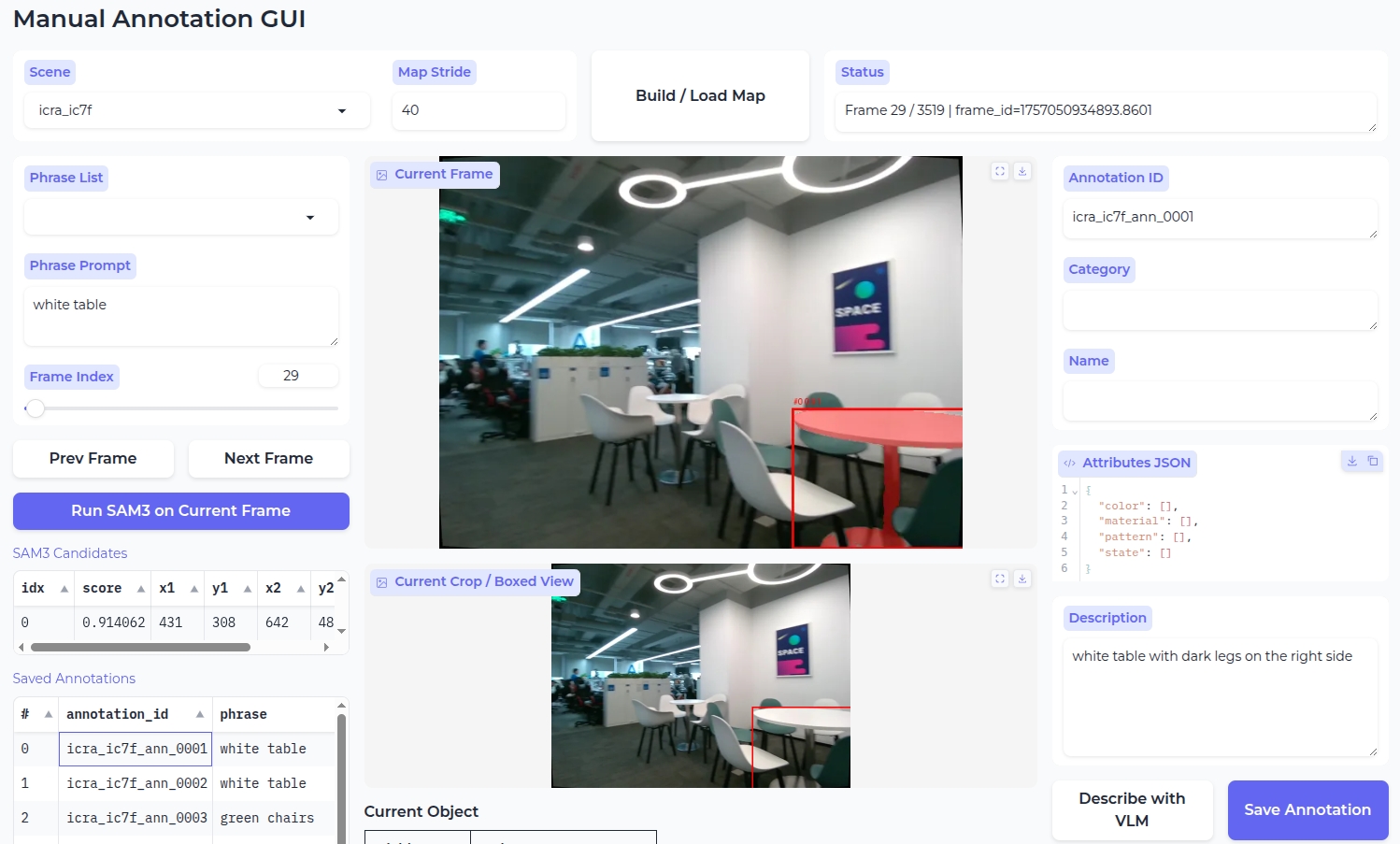}
\end{minipage}
\caption{Annotation and verification interface for real-world data.}
\label{fig:annotation_tool}
\end{figure}

\begin{figure}[h]
\centering
\begin{minipage}{0.48\linewidth}
\centering
\includegraphics[width=\linewidth]{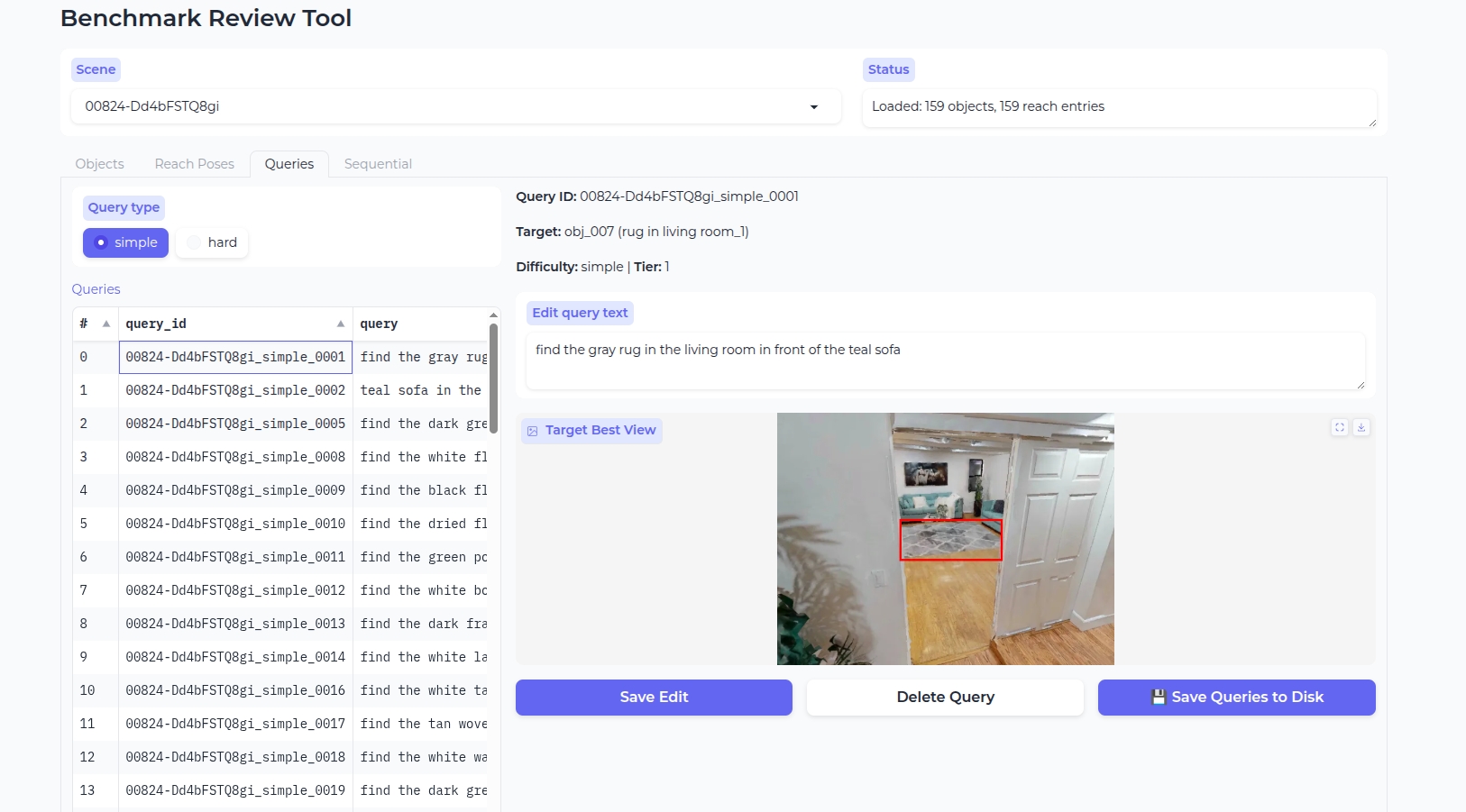}
\end{minipage}
\hfill
\begin{minipage}{0.48\linewidth}
\centering
\includegraphics[width=\linewidth]{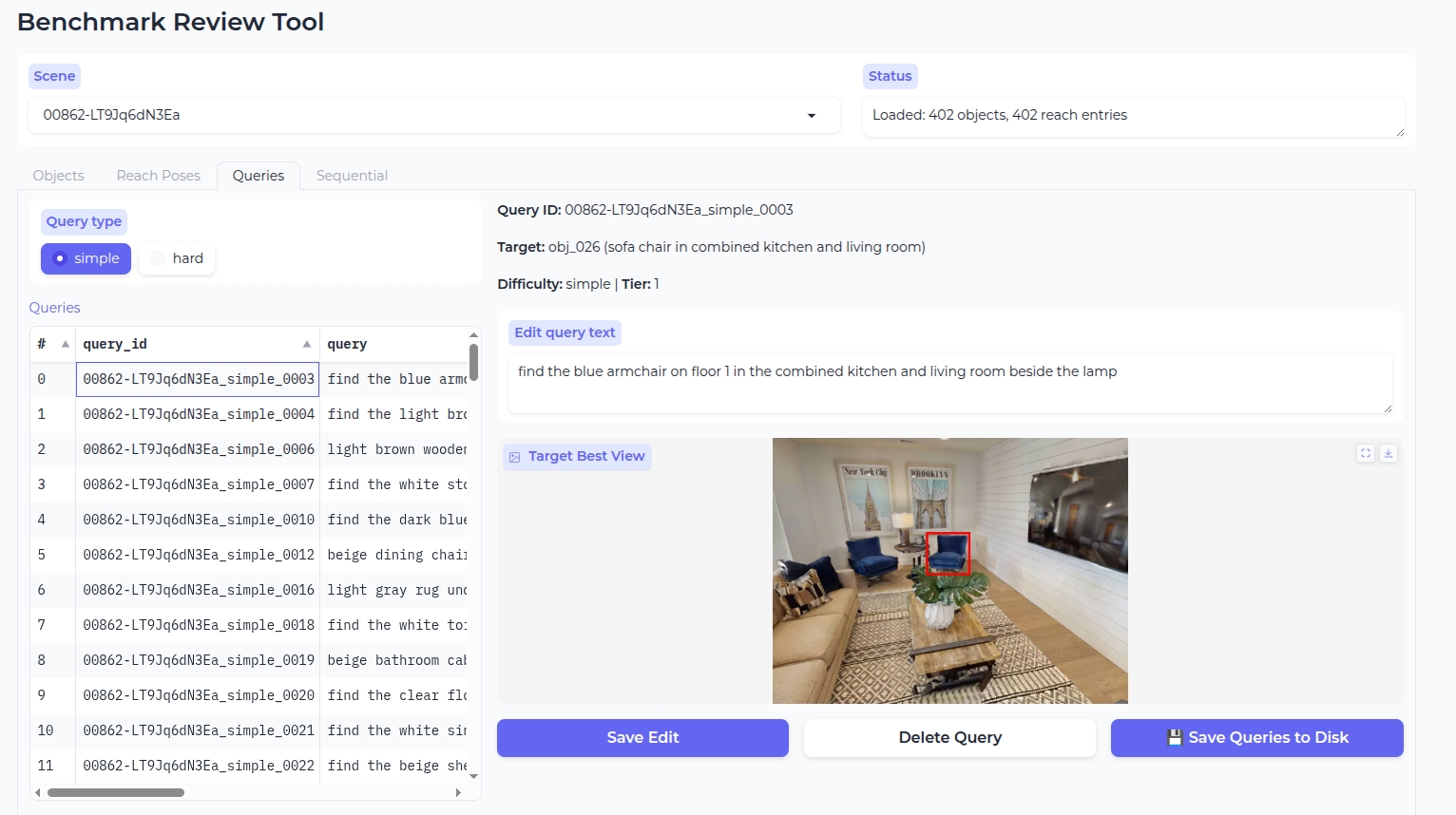}
\end{minipage}
\caption{Annotation and verification interface for simulated data.}
\label{fig:annotation_tool_sim}
\end{figure}

The annotation interfaces let annotators bind language queries to target objects, inspect visual evidence, and verify the saved 3D target metadata.
Figures~\ref{fig:annotation_tool} and~\ref{fig:annotation_tool_sim} show the real-world and simulated annotation tools.

\section{Supplementary Experiments}
\label{app:experiments}

\subsection{Evaluation protocol and metrics}
\label{app:retrieval_protocol}

Object Retrieval is evaluated separately on simulated and real-world scenes and then combined using query-weighted averaging.
The simulated benchmark contains both simple and hard queries; the real-world benchmark currently contains only hard queries.
Success@$r$ requires both a category match and centroid distance below radius $r$; LocErr is the mean centroid distance over Success@1m cases.
For ambiguous same-category queries, a prediction succeeds if it falls within the radius threshold of any valid same-category ground-truth instance under the query scope.
Sequential grounding and spatio-temporal QA follow the metrics summarized in Table~\ref{tab:metrics}.

\subsection{Per-scene retrieval breakdowns}
\label{app:per_scene_retrieval}

\begin{table}[!t]
\centering
\setlength{\abovecaptionskip}{2pt}
\setlength{\belowcaptionskip}{3pt}
\setlength{\defaultaddspace}{1.5pt}
\renewcommand{\arraystretch}{0.92}
\caption{Per-scene Object Retrieval results on simulated scenes, split by hard, simple, and combined protocols. Values are percentages for S@0.5 and S@1; Err is mean distance in meters over S@1 successes.}
\label{tab:app_sim_scene_results}
\scriptsize
\resizebox{\linewidth}{!}{%
\begin{tabular}{llr ccc ccc ccc ccc}
\toprule
& & & \multicolumn{3}{c}{HOV-SG} & \multicolumn{3}{c}{DualMap} & \multicolumn{3}{c}{FSR-VLN} & \multicolumn{3}{c}{Finder} \\
\cmidrule(lr){4-6} \cmidrule(lr){7-9} \cmidrule(lr){10-12} \cmidrule(l){13-15}
Scene & Split & Num & S@0.5 & S@1 & Err & S@0.5 & S@1 & Err & S@0.5 & S@1 & Err & S@0.5 & S@1 & Err \\
\midrule
00824-Dd4bFSTQ8gi & Hard & 91 & 20.88 & 39.56 & 0.504 & 25.27 & 43.96 & 0.424 & 36.26 & 51.65 & 0.420 & 61.54 & 71.43 & 0.222 \\
00824-Dd4bFSTQ8gi & Simple & 101 & 32.67 & 53.47 & 0.397 & 31.68 & 49.50 & 0.382 & 37.62 & 62.38 & 0.424 & 58.42 & 69.31 & 0.255 \\
00824-Dd4bFSTQ8gi & Overall & 192 & 27.08 & 46.88 & 0.440 & 28.64 & 46.87 & 0.402 & 36.98 & 57.29 & 0.422 & 59.90 & 70.31 & 0.239 \\
\addlinespace
00829-QaLdnwvtxbs & Hard & 55 & 18.18 & 32.73 & 0.404 & 23.64 & 34.55 & 0.322 & 30.91 & 45.45 & 0.449 & 49.09 & 54.55 & 0.187 \\
00829-QaLdnwvtxbs & Simple & 67 & 25.37 & 41.79 & 0.466 & 38.81 & 50.75 & 0.289 & 50.75 & 68.66 & 0.366 & 53.73 & 59.70 & 0.215 \\
00829-QaLdnwvtxbs & Overall & 122 & 22.13 & 37.71 & 0.442 & 31.97 & 43.45 & 0.304 & 41.81 & 58.20 & 0.395 & 51.64 & 57.38 & 0.203 \\
\addlinespace
00843-DYehNKdT76V & Hard & 52 & 32.69 & 48.08 & 0.454 & 38.46 & 55.77 & 0.398 & 50.00 & 59.62 & 0.324 & 53.85 & 57.69 & 0.151 \\
00843-DYehNKdT76V & Simple & 66 & 42.42 & 62.12 & 0.397 & 33.33 & 46.97 & 0.385 & 54.55 & 66.67 & 0.328 & 63.64 & 66.67 & 0.155 \\
00843-DYehNKdT76V & Overall & 118 & 38.13 & 55.93 & 0.418 & 35.59 & 50.85 & 0.391 & 52.54 & 63.56 & 0.326 & 59.32 & 62.71 & 0.153 \\
\addlinespace
00847-bCPU9suPUw9 & Hard & 78 & 19.23 & 42.31 & 0.546 & 35.90 & 44.87 & 0.292 & 15.38 & 35.90 & 0.557 & 71.79 & 74.36 & 0.157 \\
00847-bCPU9suPUw9 & Simple & 83 & 44.58 & 62.65 & 0.386 & 40.96 & 54.22 & 0.306 & 38.55 & 62.65 & 0.430 & 66.27 & 73.49 & 0.198 \\
00847-bCPU9suPUw9 & Overall & 161 & 32.30 & 52.80 & 0.448 & 38.51 & 49.69 & 0.299 & 27.32 & 49.69 & 0.475 & 68.94 & 73.91 & 0.178 \\
\addlinespace
00861-GLAQ4DNUx5U & Hard & 69 & 17.39 & 33.33 & 0.471 & 44.93 & 53.62 & 0.298 & 20.29 & 31.88 & 0.468 & 53.62 & 55.07 & 0.160 \\
00861-GLAQ4DNUx5U & Simple & 89 & 42.70 & 58.43 & 0.359 & 44.94 & 59.55 & 0.355 & 34.83 & 50.56 & 0.430 & 59.55 & 66.29 & 0.187 \\
00861-GLAQ4DNUx5U & Overall & 158 & 31.65 & 47.47 & 0.393 & 44.94 & 56.96 & 0.330 & 28.48 & 42.40 & 0.443 & 56.96 & 61.39 & 0.176 \\
\addlinespace
00862-LT9Jq6dN3Ea & Hard & 138 & 14.49 & 26.09 & 0.482 & 30.43 & 39.86 & 0.322 & 14.49 & 22.46 & 0.489 & 44.20 & 50.72 & 0.221 \\
00862-LT9Jq6dN3Ea & Simple & 148 & 31.08 & 47.97 & 0.416 & 41.22 & 55.41 & 0.347 & 29.73 & 45.27 & 0.450 & 66.22 & 72.30 & 0.183 \\
00862-LT9Jq6dN3Ea & Overall & 286 & 23.08 & 37.41 & 0.439 & 36.01 & 47.91 & 0.335 & 22.38 & 34.26 & 0.462 & 55.59 & 61.89 & 0.198 \\
\addlinespace
00873-bxsVRursffK & Hard & 90 & 47.78 & 63.33 & 0.332 & 50.00 & 57.78 & 0.242 & 26.67 & 40.00 & 0.465 & 72.22 & 76.67 & 0.146 \\
00873-bxsVRursffK & Simple & 87 & 58.62 & 74.71 & 0.293 & 50.57 & 63.22 & 0.335 & 43.68 & 56.32 & 0.320 & 65.52 & 68.97 & 0.151 \\
00873-bxsVRursffK & Overall & 177 & 53.11 & 68.92 & 0.311 & 50.28 & 60.45 & 0.288 & 35.03 & 48.02 & 0.382 & 68.93 & 72.88 & 0.148 \\
\addlinespace
00877-4ok3usBNeis & Hard & 71 & 38.03 & 61.97 & 0.454 & 32.39 & 53.52 & 0.397 & 19.72 & 40.85 & 0.526 & 63.38 & 71.83 & 0.197 \\
00877-4ok3usBNeis & Simple & 75 & 54.67 & 74.67 & 0.355 & 36.00 & 60.00 & 0.427 & 45.33 & 61.33 & 0.361 & 77.33 & 84.00 & 0.186 \\
00877-4ok3usBNeis & Overall & 146 & 46.58 & 68.49 & 0.398 & 34.24 & 56.85 & 0.412 & 32.88 & 51.37 & 0.426 & 70.55 & 78.08 & 0.191 \\
\addlinespace
00890-6s7QHgap2fW & Hard & 95 & 22.11 & 46.32 & 0.519 & 40.00 & 52.63 & 0.338 & 18.09 & 43.16 & 0.518 & 56.84 & 61.05 & 0.169 \\
00890-6s7QHgap2fW & Simple & 95 & 38.95 & 56.84 & 0.371 & 46.32 & 57.89 & 0.294 & 44.21 & 58.95 & 0.357 & 65.26 & 73.68 & 0.192 \\
00890-6s7QHgap2fW & Overall & 190 & 30.53 & 51.58 & 0.438 & 43.16 & 55.26 & 0.316 & 31.15 & 51.06 & 0.426 & 61.05 & 67.37 & 0.181 \\
\addlinespace
\midrule
Query Weighted & Hard & 739 & 24.90 & 42.76 & 0.458 & 35.59 & 48.04 & 0.334 & 23.95 & 39.24 & 0.466 & 58.05 & 63.46 & 0.182 \\
Query Weighted & Simple & 811 & 40.44 & 58.32 & 0.377 & 40.69 & 55.49 & 0.346 & 40.57 & 57.71 & 0.389 & 64.12 & 70.78 & 0.192 \\
Query Weighted & Overall & 1550 & 33.03 & 50.90 & 0.409 & 38.26 & 51.94 & 0.340 & 32.65 & 48.90 & 0.419 & 61.23 & 67.29 & 0.187 \\
\bottomrule
\end{tabular}}
\end{table}

Tables~\ref{tab:app_sim_scene_results} and~\ref{tab:app_real_scene_results} provide the per-scene retrieval breakdown behind the aggregate Object Retrieval table.
We report hard and simple splits separately when both are available.
Entries marked ``--'' indicate undefined metrics, such as LocErr when a method has no S@1 successes.

\begin{table}[!t]
\centering
\setlength{\abovecaptionskip}{2pt}
\setlength{\belowcaptionskip}{3pt}
\renewcommand{\arraystretch}{0.94}
\caption{Per-scene Object Retrieval results on real-world scenes. The real benchmark currently contains only hard queries, so no simple split is reported. Values are percentages for S@0.5 and S@1; Err is mean distance in meters over S@1 successes.}
\label{tab:app_real_scene_results}
\scriptsize
\resizebox{\linewidth}{!}{%
\begin{tabular}{lr ccc ccc ccc ccc}
\toprule
& & \multicolumn{3}{c}{HOV-SG} & \multicolumn{3}{c}{DualMap} & \multicolumn{3}{c}{FSR-VLN} & \multicolumn{3}{c}{Finder} \\
\cmidrule(lr){3-5} \cmidrule(lr){6-8} \cmidrule(lr){9-11} \cmidrule(l){12-14}
Scene & Num & S@0.5 & S@1 & Err & S@0.5 & S@1 & Err & S@0.5 & S@1 & Err & S@0.5 & S@1 & Err \\
\midrule
icra\_ic3f & 18 & 33.33 & 38.89 & 0.330 & 16.67 & 27.78 & 0.345 & 11.11 & 16.67 & 0.497 & 44.44 & 50.00 & 0.326 \\
icra\_ic4f & 16 & 31.25 & 43.75 & 0.453 & 25.00 & 37.50 & 0.445 & 18.75 & 43.75 & 0.633 & 50.00 & 62.50 & 0.306 \\
icra\_ic7f & 30 & 26.67 & 36.67 & 0.434 & 3.33 & 16.67 & 0.610 & 10.00 & 23.33 & 0.492 & 20.00 & 36.67 & 0.409 \\
icra\_sh3f & 18 & 38.89 & 55.56 & 0.343 & 0.00 & 0.00 & -- & 11.11 & 16.67 & 0.425 & 38.89 & 44.44 & 0.300 \\
\midrule
Query Weighted & 82 & 31.71 & 42.68 & 0.391 & 9.76 & 19.51 & 0.494 & 12.20 & 24.39 & 0.532 & 35.37 & 46.34 & 0.339 \\
\bottomrule
\end{tabular}}
\end{table}

\subsection{Object-type response analysis}
\label{app:longtail}

\begin{table}[!t]
\centering
\setlength{\abovecaptionskip}{2pt}
\setlength{\belowcaptionskip}{3pt}
\renewcommand{\arraystretch}{0.94}
\caption{Object-type response analysis for Finder on the aggregated Object Retrieval benchmark. Queries are grouped by target-object profile rather than dataset split.}
\label{tab:category_breakdown}
\small
\begin{tabular*}{\linewidth}{@{\extracolsep{\fill}}l l c cccc@{}}
\toprule
Type & Examples & Queries & S@0.5$\uparrow$ & S@1$\uparrow$ & S@3$\uparrow$ & $T_q\downarrow$ \\
\midrule
Large fixed        & bed, sofa, cabinet, table             & 836 & 59.45 & 65.91 & 81.10 & 80.5s \\
Small clutter      & cup, bottle, vase, toy                & 368 & 55.43 & 62.23 & 70.11 & 96.1s \\
Thin / flat        & book, plate, tray, wall art           & 201 & 64.68 & 67.16 & 74.63 & 86.0s \\
Soft items         & pillow, towel, bag, clothing          & 227 & 64.76 & 73.13 & 82.38 & 92.8s \\
\bottomrule
\end{tabular*}
\end{table}

Beyond module-wise ablations, it is useful to understand which kinds of objects Finder responds to most reliably and how response time varies by target type.
Table~\ref{tab:category_breakdown} groups Object Retrieval queries by coarse target-object profile using the ground-truth target category.
The success columns use the same geometry-only radius thresholds as the main Object Retrieval table, and $T_q$ reports mean per-query runtime.

\section{Prompt Templates}
\label{app:key_prompts}

\subsection{Request-Planning Prompt}
\begin{tcolorbox}[promptbox, title=Request-Planning Prompt]
\begin{lstlisting}[style=prompt]
You are the request planner for an object-finding loop.

Your job is NOT to directly choose a final object. Your job is to decide what
this loop should search for and which contextual objects should support the
search.

You must produce:
1. the PRIMARY target object for this loop
2. zero or more SUPPORTING targets that provide memory anchors, spatial
   disambiguation, or additional context
3. search phrases for the primary target and any supporting targets that need
   active detection
4. evidence goals for this loop
5. an optional scope_plan describing how detection frame budget should be spent

Important design rules:
- There must be exactly ONE primary target.
- Supporting targets are optional.
- Only include supporting targets if they help locate or disambiguate the
  primary target.
- Supporting targets may use these usages:
  - "memory_anchor": look up or anchor from memory or metadata
  - "disambiguate": help compare candidates using spatial relations
  - "detect": actively detect this supporting object
  - "context_only": mention it in reasoning but do not actively detect it
- Every supporting target must include a support_policy with:
  - "semantic_type": one of ["anchor", "verifier", "memory"]
  - "execution_mode": one of ["global", "local_after_primary", "conditional", "none", "rescue_only"]
  - "expected_value": estimated usefulness in [0, 1]
- Search phrases must describe objects themselves, not rooms or long spatial
  descriptions.
- Every primary target must output exactly 2 short object phrases.
- Every supporting target must output exactly 2 short object phrases.
- Do not generate open-ended synonym lists or phrase expansion beyond those required counts.
- Spatial scope and frame budget policy are separate from the object phrases.
- Keep the plan compact. Do not expand the plan into unnecessary targets.
- Preserve hard constraints already known from upstream context.
- If the user query contains a relational description like "the mug near the
  coffee machine", the mug is usually primary and the coffee machine is usually
  supporting.
- If previous loop notes exist, use them carefully. They are hints, not hard
  truth.
- If `previous_loop_hints.phrase_policy.locked` is true, preserve the prior
  primary phrases and prioritize scope/frame changes over phrase changes.

\end{lstlisting}
\end{tcolorbox}
\clearpage

\subsection{Candidate Visual-Verification Prompt}
\begin{tcolorbox}[promptbox, title=Candidate Visual-Verification Prompt]
\begin{lstlisting}[style=prompt]
You are the factual visual verifier in an object-finding loop.

Your job is NOT to decide whether the loop should continue or stop.
Your job is to extract image-grounded facts for each candidate.

You will receive:
- the raw user query
- the primary target and supporting-target expectations
- a shortlist of candidate objects
- objective, relation-specific evidence for each candidate
- a manifest that maps attached images to candidate keys

Important rules:
- Base your output primarily on what is visually observable in the images.
- Use relation-specific measurements only as supporting context, not as a substitute for visual facts.
- Every attached image is a full-frame scene view. These are not cropped object-only patches.
- If an image is described as "bbox", it means the full scene frame with a box overlay, not a crop.
- Do not output loop-control fields such as decisive / continue / accept.
- Do not guess room type from the image unless it is visually obvious. Room binding will be handled separately.
- Never infer room type from the user query, candidate label, or support-target name.
- Never say a room is "assumed", "likely", or "probably" based on query context.
- If room type is not visually explicit, say "room not visually determined" or omit room claims entirely.
- If room type is unclear, say it is unclear; do not say the image lacks context when surrounding furniture, wall, floor, doorway, or nearby objects are visible.
- Keep rationale and notes factual. Do not add advice, recommendations, or loop-control guidance.
- Only cite image indices that exist in the manifest.

Allowed categorical values:
- object_type_match: strong | partial | weak | mismatch | unclear
- color_match: strong | partial | weak | mismatch | unclear
- appears_against_wall: yes | no | unclear
- wall_contact_visibility: direct | partial | not_visible
- image_context: full_room | partial_room | object_focused | unclear
- object_completeness: complete | partial | unclear

\end{lstlisting}
\end{tcolorbox}
\clearpage

\subsection{Loop-Judgment Prompt}
\begin{tcolorbox}[promptbox, title=Loop-Judgment Prompt]
\begin{lstlisting}[style=prompt]
You are the loop judge in an object-finding system.

Your job is to decide whether to:
- accept the current best candidate
- continue to another loop
- abort because the loop budget is exhausted or the evidence is too weak

You will receive:
- the user query
- factual verifier output for each candidate
- room-binding facts derived from room_info and candidate pose
- support-relation metrics
- support-detection summary
- loop history and current shortlist

Important rules:
- The verifier output is factual only. Do not expect it to decide whether the loop should continue.
- Determine room compatibility from room_info / pose facts, not from image semantics alone.
- Prefer accepting when one candidate is clearly strongest across visual facts, support relations, room binding, and loop-history stability.
- Missing confirmation is not the same as contradiction. Treat `unclear`, `ambiguous`, or absent visual confirmation as weak evidence, not negative evidence.
- Only explicit mismatches or concrete contradictory facts should block acceptance. For example, "not glass" is blocking, but "material not fully visible" is not.
- For relation queries such as beside / near / against / under / on, structured 3D support-relation metrics can be sufficient even when the support object is not co-visible in the verifier image.
- If the support target was not detected anywhere in the loop, treat relation evidence as unavailable rather than negative against a specific primary candidate.
- When support evidence is globally unavailable, only continue if the next loop will concretely change support detection strategy, such as phrase retargeting or scope/frame changes.
- If there is only one viable candidate, it already satisfies room and relation constraints, and the verifier shows no explicit contradiction, prefer accept over continue.
- If multiple top candidates all satisfy the core constraints and remain hard to distinguish, treat this as local query ambiguity rather than a failure. In that case, prefer accepting the highest-scoring candidate instead of continuing to chase uniqueness.
- Continue only when another loop is likely to change the ranking, reveal a missing competing candidate, or resolve a concrete contradiction that matters to the final decision.
- Continue only when another loop could plausibly change the answer in a concrete way.
- If the current loop already detected primary candidates, prefer scope/frame changes over phrase changes.
- If the current loop detected zero primary candidates, a continue decision may split recovery between scope/frame and phrase changes.
- If the loop budget is already exhausted, do not return continue.
- Use next_loop_hints to focus the next loop on the best candidate(s) and deprioritize weak ones.
- Keep the reason factual and concise. Do not add motivational language or speculative claims.
- Do not repeat room guesses from visual evidence when room_info / pose facts already determine room compatibility.

\end{lstlisting}
\end{tcolorbox}

\end{document}